\documentclass[10pt,twocolumn,letterpaper]{article}

\usepackage{iccv}
\usepackage{times}
\usepackage{epsfig}
\usepackage{graphicx}
\usepackage{amsmath}
\usepackage{amssymb}
\usepackage{amsthm}

\usepackage{microtype}
\usepackage[htt]{hyphenat}
\usepackage{graphicx}
\usepackage[dvipsnames]{xcolor}
\usepackage{subcaption}
\usepackage{caption}
\usepackage{enumitem}
\usepackage{multirow}
\usepackage{tablefootnote}
\usepackage{booktabs}
\usepackage{wrapfig}
\usepackage{lipsum, afterpage, refcount}

\usepackage[pagebackref=true,breaklinks=true,letterpaper=true,colorlinks,bookmarks=false]{hyperref}

\iccvfinalcopy 

\ificcvfinal\fi

\begin{document}

\title{Distilling Large Vision-Language Model with Out-of-Distribution Generalizability}

\author{
Xuanlin Li\footnotemark[1]\thanks{\,Equal contributions; corresponding authors: xul012@ucsd.edu, yuf026@ucsd.edu}\qquad
Yunhao Fang\footnotemark[1]\qquad
Minghua Liu\qquad
Zhan Ling\qquad
Zhuowen Tu\qquad
Hao Su \qquad 
\vspace{0.1cm} \\
UC San Diego
}

\maketitle
\ificcvfinal\thispagestyle{empty}\fi

\begin{abstract}
Large vision-language models have achieved outstanding performance, but their size and computational requirements make their deployment on resource-constrained devices and time-sensitive tasks impractical. Model distillation, the process of creating smaller, faster models that maintain the performance of larger models, is a promising direction towards the solution. This paper investigates the distillation of visual representations in large teacher vision-language models into lightweight student models using a small- or mid-scale dataset. Notably, this study focuses on open-vocabulary out-of-distribution (OOD) generalization, a challenging problem that has been overlooked in previous model distillation literature. We propose two principles from vision and language modality perspectives to enhance student's OOD generalization: \textbf{(1)} by better imitating teacher's visual representation space, and carefully promoting better coherence in vision-language alignment with the teacher; \textbf{(2)} by enriching the teacher's language representations with informative and finegrained semantic attributes to effectively distinguish between different labels. We propose several metrics and conduct extensive experiments to investigate their techniques. The results demonstrate significant improvements in zero-shot and few-shot student performance on open-vocabulary out-of-distribution classification, highlighting the effectiveness of our proposed approaches. Project poster: \href{https://xuanlinli17.github.io/pdfs/iccv23_large_vlm_distillation_poster.pdf}{this link}. Code: \href{https://github.com/xuanlinli17/large_vlm_distillation_ood}{this link}.

\end{abstract}

\begin{figure}[t]
\centering
\includegraphics[width=1.0\linewidth]{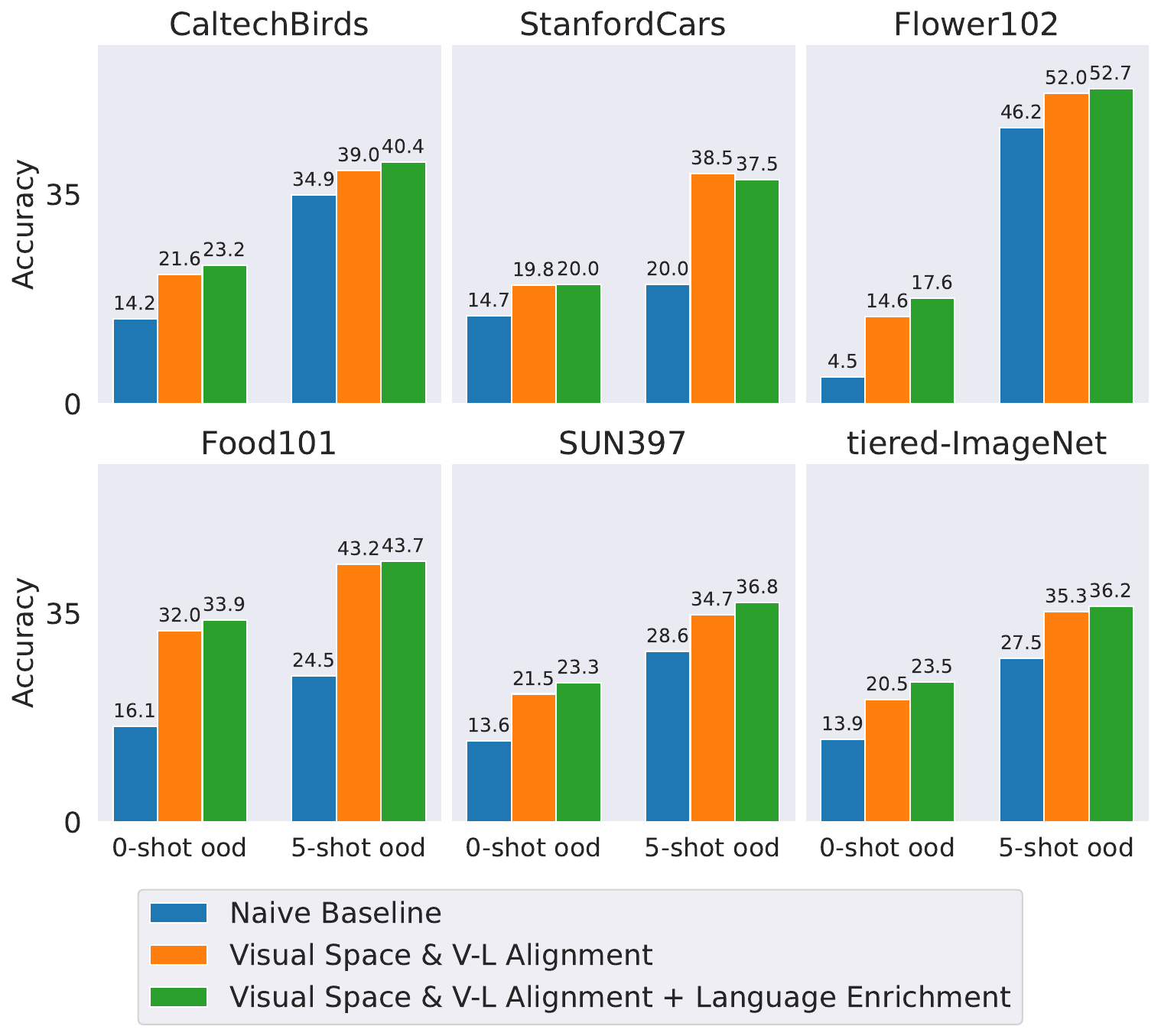}
\vspace{-1.5em}
\caption{
    Improvements of student's open-vocabulary classification accuracy on \textit{unseen} out-of-distribution (OOD) concepts across different datasets.
    We strengthen student's OOD generalization ability from two key perspectives: 1) by better imitating teacher's visual representation space, and carefully promoting better coherence in vision-language alignment with the teacher; 2) by enriching teacher’s language representations with more informative, finegrained, and meaningful attributes to effectively distinguish between different labels. The latter can be accomplished using large language models (LLMs) such as ChatGPT.
    }
    \label{fig:overall_results}
\end{figure}

\begin{figure*}[t]
\begin{minipage}[t]{0.76\textwidth}
    \centering
    \includegraphics[width=1.0\linewidth]{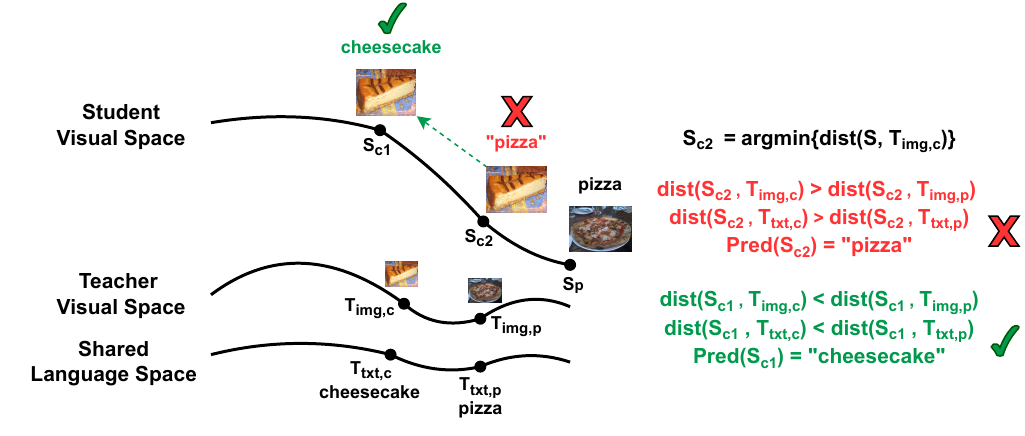}
    \vspace{-1em}
    \subcaption{}
    \label{fig:visualization_teaser_vspace}
\end{minipage}
\hspace{0.02\textwidth}
\begin{minipage}[t]{0.21\textwidth}
    \centering
    \includegraphics[width=1.0\linewidth]{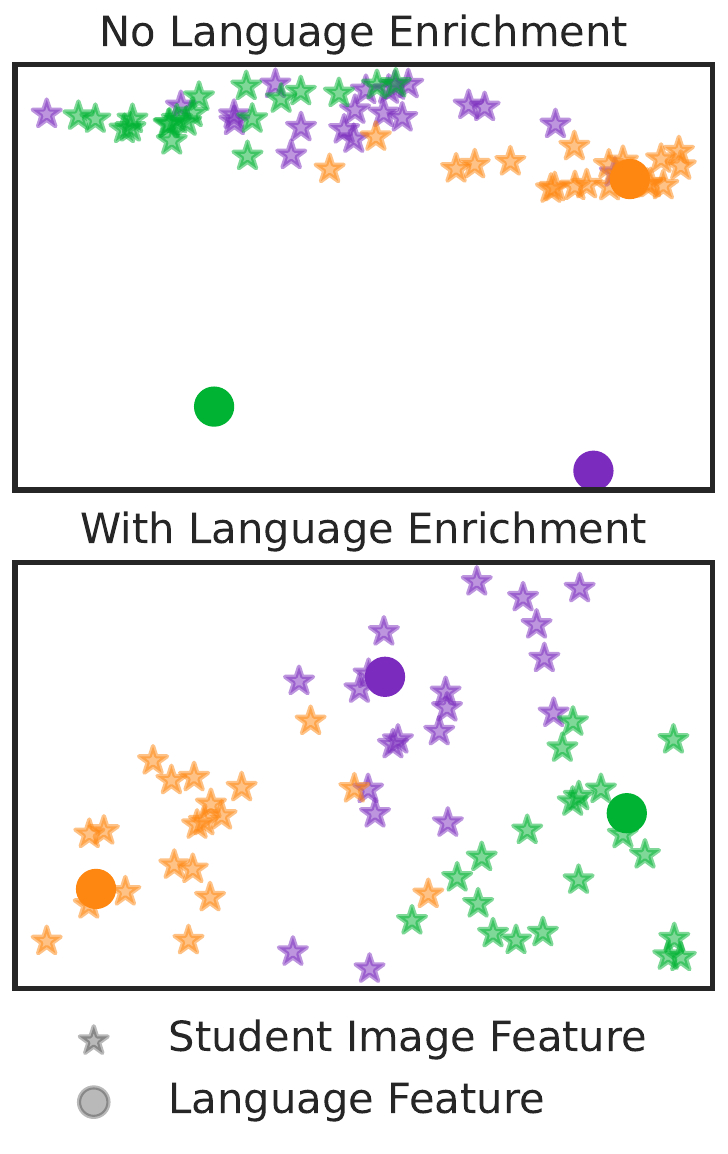}
    \vspace{-1em}
    \subcaption{}
    \label{fig:visualization_teaser_umap}
\end{minipage}
\caption{\textbf{(a)} Illustration of better teacher-student visual space alignments, motivated by our finding that achieving precise matching between teacher and student’s high-dimensional visual spaces is inherently challenging. While the student cheesecake image feature $S_{\textrm{c2}}$ minimizes its distance to the teacher's cheesecake image feature, it is even closer to the teacher's pizza image feature. This discrepancy results in poor coherence in visual space and vision-language alignment with the teacher, causing classification mistakes. By replacing the ``minimizing distance'' requirement with the ``relative distance'' requirement, which encourages student visual features to be closer to their corresponding teacher visual features than other teacher visual features, we greatly enhance visual space and vision-language alignment coherency with the teacher, thereby improving OOD generalization.
\textbf{(b)} UMAP embeddings of student visual features and text features before and after language representation enrichment with LLMs. Different colors denote different OOD classes on tiered-ImageNet. Language representation enrichment confers better clusterings of image features around their corresponding text features, enhancing OOD generalization.}
\label{fig:visualization_teaser}
\end{figure*}

\section{Introduction}

In recent years, there has been a significant growth and development of large vision language models (large VLMs) such as CLIP~\cite{clip2021}, GLIP~\cite{glip2022}, OFA~\cite{ofa2022}, SimVLM~\cite{simvlm2021}, BEiT~\cite{beit2022}, Florence~\cite{yuan2021florence}, and Flamingo~\cite{alayrac2022flamingo}, which have been pretrained on massive amounts of internet-scale data. These models have demonstrated enormous potential in a wide range of downstream applications, including classification/detection, image-to-text generation, and vision-language reasoning, especially for the out-of-distribution (OOD) samples and open-vocabulary settings. Despite their potential, the large model sizes, high computational resource requirements, and inefficient inference speed of these models restrict their deployment on mobile and IoT edge devices~\cite{howard2017mobilenets,tan2019efficientnet}, as well as in scenarios like robotic control~\cite{brohan2022rt} that require rapid network inference.  Thus, it would be ideal if we can obtain small and compact models that still possess strong generalizability towards  diverse open-set concepts encountered in the real world.

We note that the internet-scale data has endowed foundation models with well-aligned vision and language representation spaces across diverse domains and datasets. Ideally, if such representation spaces can be perfectly distilled into smaller models, the resulting models can naturally possess similar open-set OOD generalization ability as their larger counterparts. However, it is usually not the case, as demonstrated in our later experiments. Therefore, we ask the following question: \textit{How to effectively distill representation spaces of large vision-language teacher models to benefit OOD generalization of smaller student models?}

In this study, we investigate the principles and techniques for distilling visual representations from large vision-language models using small- to mid-scale datasets, with a specific focus on out-of-distribution (OOD) generalization. Although large-scale datasets with image-text pairs exist, as a pioneer work in this direction, we argue that small- to mid-scale datasets has many practical scenarios for vision application researchers (e.g., for robotics), because this is more flexible, allowing for faster research and development cycles with fewer resource dependencies. Additionally, as we shall see, extensive experiments yield a deeper understanding of the representation spaces of vision-language models, which can be beneficial for future tasks like distilling image-based foundation models for detection~\cite{kamath2021mdetr,ofa2022}, segmentation~\cite{li2022language,ding2022open}, and other data modalities like 3D geometries~\cite{xue2022ulip,peng2022openscene,liu2023openshape}. 
To keep study focused and fundamental, we also choose the image classification task to explore many possible strategies. 

Based on extensive experimental results, we propose to preserve the internal structure of the representation spaces. Specifically, we propose techniques to maintain the relationship between the visual and language representation spaces of the teacher models and a novel strategy to enhance the text representation used during distillation. Across our study, we make the following contributions:
\vspace{0.25em}\\\textbf{1)} We motivate the ability for students to generalize towards out-of-distribution (OOD) concepts by designing several metrics that measure the visual representation space consistency and the vision-language alignment consistency between the student and the teacher vision-language model.
\\\textbf{2)} We find that by better imitating teacher's visual representation space, and carefully promoting better coherence in vision-language alignment with the teacher, we substantially strengthen student's OOD generalization ability.
\\\textbf{3)} We further improve student's OOD generalization ability by enriching teacher's language representations with more informative, finegrained, and meaningful semantic attributes to effectively distinguish between different labels.
\\\textbf{4)} We conduct a thorough experimental analysis to understand the efficacy and impact of our techniques on students' OOD generalization ability.

\section{Related Work}

\noindent\textbf{Vision-Language Models (VLMs)}. Recent years have
witnessed tremendous progress on vision language models~\cite{clip2021,glip2022,ofa2022,simvlm2021,beit2022,yuan2021florence,alayrac2022flamingo}. Various lines of approaches have been proposed to improve the downstream task performance of VLMs, such as prompt learning~\cite{coop2022,yao2021cpt,zang2022unified,zhou2022learning}, prompt engineering~\cite{pratt2022platypus,hertz2022prompt,liu2022design}, and finetuning via vision-language fusion~\cite{shen2021much,wang2022image,li2023blip}. Different from these works that utilize existing VLMs on downstream tasks, our work studies distilling large VLM's feature space structures towards smaller student networks. Specifically, we investigate principles that effectively leverage the visual representation spaces and vision-language alignments in large VLMs to enhance the OOD generalization ability of students.

\noindent\textbf{Network Distillation on VLMs}. The teacher-student distillation framework~\cite{distillhinton} aims to distill teacher model knowledge into student models, thereby improving their downstream task performance. A classic setting considers teachers and students to be vision-only. In particular, distilling image representations from teacher visual backbones, such as CNNs~\cite{resnet2016} and Vision Transformers~\cite{vit2021}, have been well-studied~\cite{deit2021,park2019relational,liu2019knowledge,chen2021distillingfeaturereview,tian2019contrastive,peng2022unified,fang2021seed,wu2022tinyvit}. With the success of large-scale VLMs, recent works start to distill their powerful multimodal visual and language representations towards various downstream tasks, such as object detection~\cite{vild2021,ma2022open}, visual-language reasoning~\cite{cliptd2022,multimodelgeneration2022,fang2021compressing}, scene understanding~\cite{peng2022openscene}, or adopt distillation during vision language model pretraining to further enhance their performance ~\cite{blip2022,li2021align,dong2022maskclip,zhong2022regionclip,tian2021vl}. 

As the real world is full of concepts unseen during model training, it is essential for student models to attain strong OOD generalization capabilities. However, prior works have rarely \textit{carefully} studied how to design distillation approaches to enhance such ability. In this work, we propose principles and approaches to improve student's OOD generalization ability, and we carefully analyze their efficacies through various metrics and extensive experiments. 
\section{Overview}
\label{sec:overview}

\noindent\textbf{Problem Setup}. We distill a large vision-language teacher model $T$ (e.g., CLIP ViT-L/14) to a small student image model $S$ (e.g., ResNet18) by focusing on out-of-distribution (OOD) generalization for open-vocabulary object classification. We choose small- or mid-scale datasets to achieve the distillation so that the distillation process is flexible for fast research cycle and has less resource dependency. 

The teacher consists of an image encoder $T_{\textrm{img}}(\cdot)$ and a text encoder $T_{\textrm{txt}}(\cdot)$.
During distillation, we keep the flexibility of the existing teacher text encoder for the open-set setting, and we let the student model $S$ be vision-only, i.e., $S=S_{\textrm{img}}$.
Through this process, we hope that $S$ not only achieves high prediction accuracy on seen labels $\mathcal{Y}_\mathrm{id}$, but also attains strong generalization ability on out-of-distribution labels $\mathcal{Y}_\mathrm{ood}$. In addition, we train students from scratch to avoid label contamination, allowing us to more carefully assess and understand their OOD generalization ability.

 \noindent\textbf{Experiment Setup}. We are given a training (distillation) dataset $\mathcal{X}_{\textrm{train}}$, an in-distribution evaluation dataset $\mathcal{X}_{\textrm{id}}$, and an out-of-distribution evaluation dataset $\mathcal{X}_{\textrm{ood}}$. Each dataset consists of image-label pairs $\{(\mathbf{x}, y)\}$. For $(\mathbf{x}, y) \in \mathcal{X}_{\textrm{train}} \cup \mathcal{X}_{\textrm{id}}$,  we have $y \in \mathcal{Y}_\mathrm{id}$, and for $(\mathbf{x}, y) \in \mathcal{X}_{\textrm{ood}}$, we have $y \in \mathcal{Y}_\mathrm{ood}$, where $\mathcal{Y}_\mathrm{id}, \mathcal{Y}_\mathrm{ood}$ denote in-distribution and out-of-distribution label sets, and $\mathcal{Y}_\mathrm{id}\cap \mathcal{Y}_\mathrm{ood}=\varnothing$. We evaluate the student on $\mathcal{X}_{\textrm{id}}$, along with zero-shot and few-shot generalization on $\mathcal{X}_{\textrm{ood}}$. For few-shot learning, we finetune our student models (we will show later in our ablation study that finetuning achieves much higher performance on $\mathcal{X}_{\textrm{ood}}$ than training-free retrieval). We also adopt balanced training batches where at most half of samples come from few-shot data and the rest from $\mathcal{X}_{\textrm{train}}$. This strategy leads to similar performance on $\mathcal{X}_{\textrm{id}}$ before and after finetuning. More implementation details are presented in Appendix~\ref{sec:app_hyperparams}.
 
 We adopt a diverse collection of recognition tasks using small to medium-scale datasets, including CaltechBirds~\cite{caltechbirds2010},  StanfordCars~\cite{stanfordcars}, Flower102~\cite{flower102}, Food101~\cite{food101}, SUN397~\cite{sun397}, and tiered-ImageNet~\cite{tiered2018,imagenet2009}. We split the dataset labels such that $|\mathcal{Y}_\mathrm{id}|=|\mathcal{Y}_\mathrm{ood}|$, except tiered-ImageNet, which comes with an existing split.
Detailed dataset statistics are listed in Appendix~\ref{sec:app_dataset_stats}.

\noindent\textbf{Organization of Exposition}. Like other papers on understanding neural network behaviors~\cite{2017calibration,2017memorization,henderson2018deep,wang2021understanding}, we will propose techniques and invent metrics to help understand the effectiveness of each technique. Following the organization of a representative work of this kind~\cite{zhang2017understanding}, we present different techniques in a waterfall style and organize the presentation of each technique by grouping the approach, metric, and results together.

\noindent\textbf{Approach Overview}. Our approach aims to preserve the structure of the visual representation space and its relationship with the text representation space inherited from the teacher model to enhance the student's OOD generalization ability. Specificially, \textbf{(1)} In Sec.~\ref{sec:only_image}, we show that by better imitating teacher's visual representation space, and carefully promoting better coherence in vision-language alignment with the teacher, students achieve substantially better OOD generalization. \textbf{(2)} In Sec.~\ref{sec:img_and_text}, we show that by enriching teacher's language representations, there are more meaningful attributes to effectively distinguish between different labels, thereby further enhancing student's OOD generalization ability. A summary of our main experimental findings is illustrated in Fig.~\ref{fig:overall_results} and Fig.~\ref{fig:visualization_teaser}.
\section{Teacher-Student Visual Space and Vision-Language Alignments}
\label{sec:only_image}

\begin{table*}
\centering
\scriptsize
\setlength{\tabcolsep}{4.0pt}
\begin{tabular}{l|cccccc|c} 
    \toprule
    & CaltechBirds & StanfordCars & Flower102 & Food101 & SUN397 & tiered-ImageNet & \textbf{Average}\\ \midrule
    {$\text{CLIP ViT-L/14}$} & 70.0~/~70.5 & 79.3~/~78.3 & 74.4~/~84.1 & 90.5~/~91.2 & 72.8~/~74.4 & 71.1~/~76.3 & 76.4~/~79.1\\    
    {$\text{CLIP RN50}$} & 57.1~/~56.4 & 53.2~/~56.3 & 59.3~/~65.3 & 76.5~/~78.3 & 65.0~/~66.3 & 55.7~/~62.0 & 61.1~/~64.1 \\ \midrule 
    {$\text{Closed-Set Classification}$} & 48.1~/~NA~/~18.1 & 27.9~/~NA~/~10.1 & 77.1~/~NA~/~45.0 & 71.7~/~NA~/~30.3 & 57.8~/~NA~/~31.1 & 63.4~/~NA~/~31.2 & 57.7~/~NA~/~27.6\\
    {$\mathcal{L}_{\textrm{cls}}$} & 61.0~/~14.2~/~34.9 & 56.3~/~14.7~/~20.0 & 81.2~/~4.5~/~46.2 & 72.2~/~16.1~/~24.5 & 57.5~/~13.6~/~28.6 & 64.4~/~13.9~/~27.5 & 65.4~/~12.8~/~30.3\\
    {$\mathcal{L}_{\textrm{mse}}$}  & 27.0~/~12.0~/~14.3 & 5.5~/~3.8~/~4.0 & 48.1~/~7.3~/~15.0 & 45.0~/~17.0~/~19.3 & 24.3~/~11.0~/~14.5 & 49.3~/~14.8~/~23.2 & 33.2~/~11.0~/~15.1\\
    {$\mathcal{L}_{\textrm{cls}} + \mathcal{L}_{\textrm{mse}}$} & \textbf{63.7}~/~17.4~/~36.2 & 62.2~/~18.8~/~35.1 & 82.6~/~6.3~/~46.0 & 72.3~/~19.0~/~35.5 & 57.1~/~15.3~/~29.4 & 66.2~/~14.9~/~28.5 & 67.4~/~15.3~/~35.1\\
    {$\mathcal{L}_{\textrm{im-cst}}$} & 42.1~/~21.3~/~29.1 & 33.2~/~13.7~/~20.0 & 54.8~/~13.3~/~27.3 & 70.0~/~34.9~/~36.8 & 45.2~/~22.8~/~27.2 & 46.3~/~22.8~/~30.8 & 48.6~/~21.5~/~28.5\\
    {$\mathcal{L}_{\textrm{cls}} + \mathcal{L}_{\textrm{im-cst}}$} & 60.9~/~20.4~/~37.6 & 59.6~/~18.3~/~31.2 & 82.4~/~12.7~/~\textbf{52.5} & 74.0~/~30.5~/~42.0 & \textbf{62.5}~/~18.8~/~\textbf{35.2} & 64.4~/~18.0~/~33.5 & 67.3~/~19.8~/~38.7\\
    {$\mathcal{L}_{\textrm{cls}} + \mathcal{L}_{\textrm{im-cst}} + \mathcal{L}_{\textrm{mse}}$} & 62.5~/~20.8~/~\textbf{39.0} & 59.6~/~19.0~/~33.1 & 82.6~/~12.0~/~48.7 & \textbf{75.0}~/~31.2~/~42.0 & 60.0~/~19.8~/~\textbf{35.2} & 67.0~/~19.4~/~34.6 & 67.8~/~20.3~/~\textbf{38.8}\\
    {$\mathcal{L}_{\textrm{cls}} + \mathcal{L}_{\textrm{im-cst}} + \mathcal{L}_{\textrm{vlprox}} (+ \mathcal{L}_{\textrm{mse}})$\footnotemark} & 62.3~/~21.6~/~\textbf{39.0} & \textbf{63.9}~/~\textbf{19.8}~/~38.5 & \textbf{82.7}~/~\textbf{14.6}~/~52.0 & 74.3~/~32.0~/~\textbf{43.2} & 61.7~/~21.5~/~34.7 & \textbf{67.5}~/~20.5~/~\textbf{35.3} & \textbf{68.7}~/~21.7~/~\textbf{40.5}\\
    $\mathcal{L}_{\textrm{im-cst}} + \mathcal{L}_{\textrm{vlprox}} (+ \mathcal{L}_{\textrm{mse}})$ & 45.3~/~\textbf{21.9}~/~30.4 & 46.5~/~17.8~/~26.9 & 66.9~/~13.5~/~35.4 & 71.4~/~\textbf{35.2}~/~40.0 & 52.0~/~\textbf{23.1}~/~28.8 & 57.5~/~\textbf{23.0}~/~33.2 & 56.6~/~\textbf{22.4}~/~32.5\\
    \bottomrule
\end{tabular}
\caption{Comparison between student models trained without teacher-student visual representation space alignment ($\mathcal{L}_{\textrm{cls}}$ only), with direct teacher visual feature fitting ($+\mathcal{L}_{\textrm{mse}}$), 
with improved teacher-student visual space alignment ($+\mathcal{L}_{\textrm{im-cst}}$), and with improved preservation of teacher's vision-language alignment structure ($+\mathcal{L}_{\textrm{vlprox}}$). We adopt ResNet18 as the student and CLIP ViT-L/14 as the teacher. The three numbers $x_1/x_2/x_3$ in each entry denote the evaluation performance on $\mathcal{X}_{\textrm{id}}$, zero-shot performance on $\mathcal{X}_{\textrm{ood}}$, and 5-shot performance on $\mathcal{X}_{\textrm{ood}}$, respectively (``NA''=not applicable). As reference, we also report CLIP performance ($x_1/x_2$ in each entry denote zero-shot results on $\mathcal{X}_{\textrm{id}}$ and $\mathcal{X}_{\textrm{ood}}$), along with a closed-set classfication baseline that uses separate classifiers for $\mathcal{Y}_\mathrm{id}$ and $\mathcal{Y}_\mathrm{ood}$. Note that CLIP was pretrained on LAION~\cite{schuhmann2021laion}, where the training image-text pairs contain many concepts similar to those in $\mathcal{Y}_\mathrm{id}$ and $\mathcal{Y}_\mathrm{ood}$, resulting in higher evaluation performance on $\mathcal{X}_\mathrm{ood}$.}
\afterpage{\footnotetext{If $\mathcal{L}_{\textrm{mse}}$ is under a parenthesis, then we report the better performance between adding and not adding $\mathcal{L}_{\textrm{mse}}$. However the impact of $\mathcal{L}_{\textrm{mse}}$ is not significant.}}
\label{tab:main_imonly}
\end{table*}

\subsection{Better Imitating Teacher's Visual Representation Space}
\label{subsec:better_visual_repr}
\noindent\textbf{Naive baseline without visual space alignment between teacher and student}. A straightforward approach to training a student model is to directly align its visual representation with teacher's language representation through \textit{only} the following contrastive loss:
\begin{equation}
\begin{aligned}
\mathcal{L}_{\textrm{cls}}(\mathbf{x}, y) &= \sum_{y'} -\mathbf{1}_{y'=y} \log P_S(y'|\mathbf{x}) \\
P_S(y|\mathbf{x}) &= \frac{\exp(\cos(S(\mathbf{x}), T_{\textrm{txt}}(l(y)))/\tau)}{\sum_{y' \in \mathcal{Y}} \exp(\cos(S(\mathbf{x}), T_{\textrm{txt}}(l(y'))) / \tau)}
\end{aligned}
\label{eqn:Lcls}
\end{equation}
Here $\tau$ is the temperature parameter. $S(\mathbf{x})$ denotes the feature of $\mathbf{x}$ extracted by a student image model $S$.  $l(y)=\textrm{prompt} + \textrm{description}(y)$ is a language generation function that maps the label $y$ to its natural language representation, which consists of a prompt and a description of $y$. For this baseline, we use the suggested prompt ``A photo of a'' from the CLIP paper, and let description($y$) to be simply the class label name (e.g., ``lotus''). From here on, we assume that all teacher and student representations are normalized, i.e., $S(\mathbf{x}) \leftarrow \frac{S(\mathbf{x})}{||S(\mathbf{x})||_2}$, $T_{\textrm{txt}}(l(y)) \leftarrow \frac{T_{\textrm{txt}}(l(y))}{||T_{\textrm{txt}}(l(y))||_2}$. In this case, $\cos(S(\mathbf{x}), T_{\textrm{txt}}(l(y))) = 1 - \frac{||S(\mathbf{x}) - T_{\textrm{txt}}(l(y))||_2^2}{2}$. 

Training student models solely using $\mathcal{L}_{\textrm{cls}}$ does not enforce similarity between teacher and student visual representation space structures. Consequently, student visual space structures tend to overfit to training labels, hindering their OOD generalizability. This motivates us to introduce auxiliary losses to align teacher and student visual representation spaces, which we will describe next.

\begin{table}[t]
\centering
\small
\begin{tabular}{l|cc}
    \toprule
    & Food101 & SUN397 \\ \midrule
    {$\mathcal{L}_{\textrm{mse}}$} & 0.24~/~28.4° & 0.36~/~34.9° \\
    {$\mathcal{L}_{\textrm{mse}}$~(RN50)} & 0.24~/~28.4° & 0.35~/~34.4° \\
    {$\mathcal{L}_{\textrm{cls}} + \mathcal{L}_{\textrm{mse}}$} & 0.45~/~39.2° & 0.71~/~49.8° \\
    {$\mathcal{L}_{\textrm{cls}}+ \mathcal{L}_{\textrm{mse}}+ \mathcal{L}_{\textrm{im-cst}}$} & 0.65~/~47.6° & 0.82~/~53.8° \\
    {$\mathcal{L}_{\textrm{cls}}+ \mathcal{L}_{\textrm{im-cst}}$} & 1.29~/~69.2° & 1.28~/~68.9° \\
    \bottomrule
\end{tabular}
\caption{Average MSE / degree difference between student visual features and teacher visual features for students trained with different strategies. We adopt ResNet18 as student and CLIP ViT-L/14 as teacher, except ``$\mathcal{L}_{\textrm{mse}}$~(RN50)'', where the student is ResNet50, and the teacher is CLIP ResNet50. We observe that it is very challenging for students to \textit{precisely} match teacher’s visual features through $\mathcal{L}_{\textrm{mse}}$. Moreover, in conjunction with Tab.~\ref{tab:main_imonly}, we observe that a lower visual feature MSE is not necessary for better student OOD generalization.}
\label{tab:mse_loss}
\end{table}

\noindent\textbf{Directly fitting teacher visual features}. We note that teacher's visual representation space is well-aligned with language across diverse datasets, and such alignment demonstrates strong generalization across many domains. By imitating teacher's visual space structure, we hope to enhance the ability for student’s visual space to generalize and extrapolate towards unseen concepts, thereby implicitly enhancing the generalizability of student's vision-language alignment and improving its OOD generalization. 

A direct approach to achieve this is to align the teacher and student visual representations through the Mean Squared-Error (MSE) loss: 
\begin{equation}
\mathcal{L}_{\textrm{mse}}(\mathbf{x}) = ||S(\mathbf{x}) - T_{\textrm{img}}(\mathbf{x})||_2^2
\end{equation}
In Tab.~\ref{tab:main_imonly}, we show that adding $\mathcal{L}_{\textrm{mse}}$ on top of $\mathcal{L}_{\textrm{cls}}$ improves student OOD generalization. However, upon further examination, we find that \textit{students face significant challenges in precisely reproducing teacher's visual representations}, as evidenced by the substantial errors shown in Tab.~\ref{tab:mse_loss}. Such errors persist even when the student and teacher networks possess the same representation power (e.g., both ResNet50 networks). This phenomenon highlights that achieving precise matching between teacher and student's high-dimensional visual feature spaces is inherently challenging, which can be attributed to differences in weight initialization, training data, and the presence of local minima in the loss landscape. Moreover, we later find that when students struggle to precisely match teacher's visual features, they also struggle to preserve teacher's \textit{local visual space structure} and \textit{relative visual feature relationship} between different images, hindering their OOD generalization ability.

\noindent\textbf{Better imitating teacher visual representation space}. Since precisely matching teacher's visual features is inherently challenging, we propose to augment the training objective with the following contrastive loss, which ``softly'' matches teacher's visual features:
\begin{equation}
    \mathcal{L}_{\textrm{im-cst}}(\mathbf{x}) = \frac{\exp(-||S(\mathbf{x}) - T_{\textrm{img}}(\mathbf{x})||_2^2/\tau)}{\sum_{\mathbf{x}'} \exp(-||S(\mathbf{x}) - T_{\textrm{img}}(\mathbf{x'})||_2^2 / \tau)}
\end{equation}
By combining $\mathcal{L}_{\textrm{im-cst}}$ with $\mathcal{L}_{\textrm{cls}}$, we observe in Tab.~\ref{tab:main_imonly} that the student exhibits significantly better zero-shot and few-shot OOD generalization ability across different datasets. Interestingly, such improvement is accompanied by a larger distance between student and teacher visual features, as shown in Tab.~\ref{tab:mse_loss}. This finding suggests that a lower MSE visual feature matching loss, which only captures the absolute distance to teacher's visual features, does \textit{not} necessarily imply better visual space consistency. This is because \textit{MSE does not consider the coherence of relative visual feature relationships or local visual space structures between teacher and student}.

In the following paragraphs, we will develop several metrics to better assess the teacher-student visual space consistency. These metrics provide us with valuable insights into how $\mathcal{L}_{\textrm{im-cst}}$ facilitates students to achieve closer visual space proximity to the teacher while yielding a deeper understanding of the teacher's visual representation space.

\begin{figure}[t]
\centering
    \includegraphics[width=1.0\linewidth]{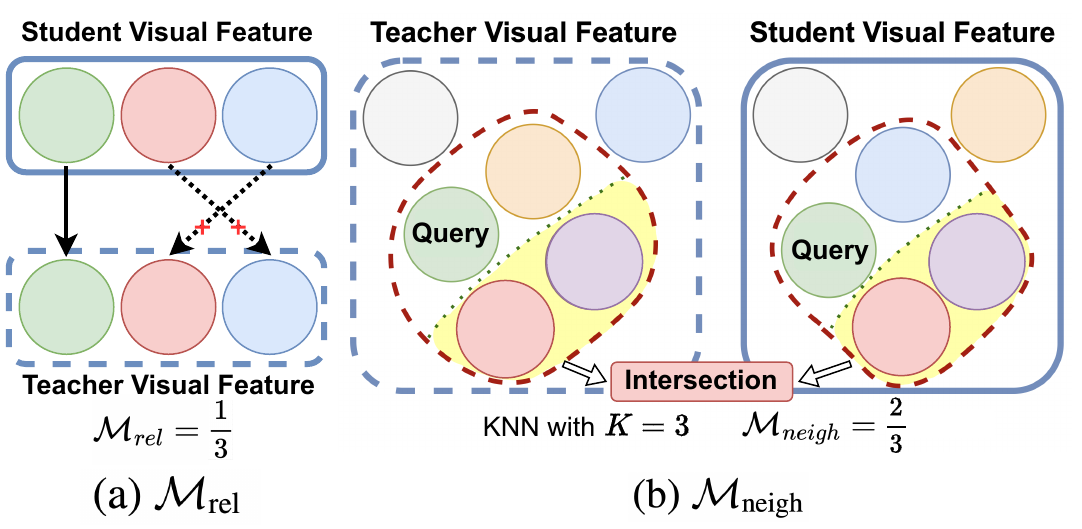}
    \caption{Illustrations of $\mathcal{M}_{\textrm{rel}}$ and $\mathcal{M}_{\textrm{neigh}}$ that measure the coherence of relative visual feature relationship and local visual space structure between the student and teacher. Each color corresponds to a single image. In (a), arrows indicate the nearest neighbors from student's visual space to teacher's visual space. Solid lines represent correct matching, while dashed lines indicate incorrect matching. In (b), the green circle represents a query image, and the red dashed regions represent the $k$-nearest neighbors (kNN) in both spaces. We measure the intersection between the two kNN sets.}
   \label{fig:M_neigh}
   \label{fig:M_rel}
\end{figure}

\noindent\textbf{Quantifying teacher-student visual space alignment}. To quantify how student models preserve teacher's visual representation space structures, we propose two metrics. For the first metric, we are motivated by the fact that if the relative relationships among teacher's visual representations are preserved, then for an image $\mathbf{x}$, $S(\mathbf{x})$ tends to be the closest to $T_{\textrm{img}}(\mathbf{x})$ rather than $T_{\textrm{img}}(\mathbf{x'})$ of some other image $\mathbf{x'} \ne \mathbf{x}$. The latter scenario tends to break the relative feature relationship between $\mathbf{x}$ and $\mathbf{x'}$ that is originally present in the teacher, thereby causing the student to have distinct visual feature manifold structure from the teacher, as illustrated in Fig.~\ref{fig:M_rel}(a). Furthermore, if $S(\mathbf{x})$ is closer to $T_{\textrm{img}}(\mathbf{x'})$, then the language feature closest to $S(\mathbf{x})$ often differs from the language feature closest to $T_{\textrm{img}}(\mathbf{x})$, and the latter is typically the ground truth language label. Thus, such discrepancy often results in erroneous label predictions. We can then formally define the first metric as follows:
\begin{equation}
    \mathcal{M}_{\textrm{rel}}(\mathcal{X}) = \frac{\sum_{i=1}^{|\mathcal{X}|} \mathbf{1}[i=\textrm{argmin}_{j} ||T_{\textrm{img}}(\mathbf{x}_j) - S(\mathbf{x}_i)||_2^2]}{|\mathcal{X}|}
\end{equation}
The second metric measures the proximity between local neighborhoods of student visual features and teacher visual features. That is, for each image, how much the $k$ images whose features have the closest proximity to it (excluding itself) overlap between the student and the teacher (see Fig.~\ref{fig:M_neigh}b for illustration). We can define this metric as follows:
\begin{equation}
    \mathcal{M}_{\textrm{neigh}}(\mathcal{X}, k) = \frac{\sum_{i=1}^{|\mathcal{X}|} |\textrm{KNN}(S, \mathbf{x}_i, k) \cap \textrm{KNN}(T_{\textrm{img}}, \mathbf{x}_i, k)|}{k|\mathcal{X}|}
\end{equation}
Here $\textrm{KNN}(S, \mathbf{x}_i, k) = \{ \textrm{argbottomk}_j  ||S(\mathbf{x}_j) - S(\mathbf{x}_i) ||_2    \}$ outputs the $k$ nearest-neighbor image-ids of $S(\mathbf{x}_i)$, and $\textrm{KNN}(T_{\textrm{img}}, \mathbf{x}_i, k)$ is defined similarly. This metric is complementary to $\mathcal{M}_{\textrm{rel}}$ as it only requires the set of neighbor image ids to be identical between the student and the teacher, and does not enforce these neighbor images to have the same relative feature relationships.

\begin{table}
\centering
\scriptsize
\begin{subtable}{0.45\textwidth}
\centering
\setlength{\tabcolsep}{3.0pt}
\begin{tabular}{l|cc|ccc}
    \toprule
    $ \mathcal{M}_{\textrm{rel}} \uparrow ~|~    \mathcal{M}_{\textrm{neigh}} \uparrow$ & $\mathcal{X}_{\textrm{train}}$ & $\mathcal{X}_{\textrm{ood}}$ & $k=3$ & $k=5$ & $k=10$ \\ \midrule
    {$\mathcal{L}_{\textrm{cls}} + \mathcal{L}_{\textrm{mse}}$} & 0.030 & 0.004 & 0.13~/~0.06 & 0.18~/~0.07 & 0.27~/~0.08\\
    {$\mathcal{L}_{\textrm{cls}}+ \mathcal{L}_{\textrm{mse}}+ \mathcal{L}_\textrm{{im-cst}}$} & 0.305 & 0.022 & 0.20~/~0.10 & 0.25~/~0.11 & 0.34~/~0.13\\
    \bottomrule
\end{tabular}
\end{subtable}
\caption{We evaluate $\mathcal{M}_{\textrm{rel}}$ (middle 2 columns) and $\mathcal{M}_{\textrm{neigh}}$ (right 3 columns) on different students to measure their coherence with teacher's relative visual feature relationships and local visual feature structures (higher the better). For $\mathcal{M}_{\textrm{neigh}}$, $x_1/x_2$ in each entry denote $\mathcal{M}_{\textrm{neigh}}(\mathcal{X}_{\textrm{train}})$ and $\mathcal{M}_{\textrm{neigh}}(\mathcal{X}_{\textrm{ood}})$, respectively. Metrics are evaluated on Flower102. More results in Appendix.}
\label{tab:metric_imonly}
\end{table}

\noindent\textbf{$\mathcal{L}_{\textrm{im-cst}}$ improves teacher-student visual space alignments through better relative and local visual space coherence}.  We assess $\mathcal{M}_{\textrm{rel}}$ and $\mathcal{M}_{\textrm{neigh}}$ on different students and present the results in Tab.~\ref{tab:metric_imonly}. We find that students trained solely with $\mathcal{L}_{\textrm{mse}}$ and without $\mathcal{L}_{\textrm{im-cst}}$ exhibit poor values for both $\mathcal{M}_{\textrm{rel}}$ and $\mathcal{M}_{\textrm{neigh}}$. Notably, $\mathcal{M}_{\textrm{rel}}(\mathcal{X}_{\textrm{train}})$ is only 0.03, indicating a large discrepancy between student and teacher visual spaces, even on the training set. On the other hand, after incorporating $\mathcal{L}_{\textrm{im-cst}}$, we observe significant improvements of $\mathcal{M}_{\textrm{rel}}$ and $\mathcal{M}_{\textrm{neigh}}$ on both $\mathcal{X}_{\textrm{train}}$ and $\mathcal{X}_{\textrm{ood}}$. This demonstrates that the student exhibits much better consistency in relative and local visual structures with the teacher, both for seen and unseen concepts, thereby enhancing the generalization and extrapolation ability of the student's visual space. A better-aligned visual space also implicitly enables better generalization and extrapolation in vision-language alignments (which we show later), contributing to improved OOD performance. Furthermore, our findings suggest a broader principle: when students face challenges in precisely matching the teacher's visual space, the proximity of local visual feature structures and relative visual feature relationships have a greater impact on OOD generalization than the absolute distance to the teacher's features.

\subsection{Better Coherence with Teacher's Vision-Language Alignment Structure}
\label{subsec:better_vl_alignment}

\noindent\textbf{Motivation.} In Section~\ref{subsec:better_visual_repr}, we focused on improving the student's OOD generalization ability by better aligning student-teacher visual spaces. Since teacher's visual space is well-aligned with language across diverse concepts and domains, a better student coherence with teacher's visual space \textit{implicitly} leads to better vision-language (V-L) alignments. Naturally, an alternative perspective to improve student's OOD generalization becomes to enhance its \textit{explicit} V-L alignments and improve their coherence with the teacher's, where we previously only used a simple contrastive V-L matching loss $\mathcal{L}_{\textrm{cls}}$. Another motivation to focus on \textit{explicit} V-L alignments arises from our finding that they play an essential role to ensure precise and accurate V-L alignments, especially when training on seen concepts or performing few-shot learning on novel concepts. \textit{Relying solely on implicit V-L alignments is inadequate in these scenarios}. This is evident in Tab.~\ref{tab:main_imonly}, where \textit{solely} utilizing the visual space alignment loss $\mathcal{L}_{\textrm{im-cst}}$ yields better performance on 0-shot $\mathcal{X}_{\textrm{ood}}$ (where classes are unseen) but worse performance on $\mathcal{X}_{\textrm{id}}$ and 5-shot $\mathcal{X}_{\textrm{ood}}$ (where classes are seen). On the other hand, by combining both implicit and explicit V-L alignment losses ($\mathcal{L}_{\textrm{im-cst}} + \mathcal{L}_{\textrm{cls}}$), students excel in all of $\mathcal{X}_{\textrm{id}}$, 0-shot $\mathcal{X}_{\textrm{ood}}$, and 5-shot $\mathcal{X}_{\textrm{ood}}$ scenarios. Therefore, by improving explicit V-L alignments, we not only hope to further enhance student's 0-shot OOD generalization ability, but also improve their performance on familiar concepts and their ability to few-shot adapt to novel concepts.

\noindent\textbf{Explicitly enhancing teacher-student vision-language alignment coherency.} We note that while $\mathcal{L}_{\textrm{cls}}$ performs explicit V-L alignments, it has the limitation of indiscriminately pushing an image away from all non ground-truth language features, therefore disregarding teacher's relative alignment relationship between the same image and different language features. Furthermore, we find that even though preserving teacher's relative V-L alignment structure is desirable, it may not always be perfect due to potential misalignments between teacher image features and their corresponding language labels. These misalignments can introduce inconsistent noise during distillation, ultimately harming student performance.

Motivated by these observations, we propose to augment our training objective with $\mathcal{L}_{\textrm{vlprox}}$, which effectively and carefully preserves the teacher's vision-language alignment structure while accounting for potential misalignments:
\begin{equation}
\begin{aligned}
\mathcal{L}_{\textrm{vlprox}}(\mathbf{x}, k) &= I(\mathbf{x}) \cdot 
  \mathcal{D}_{\textrm{KL}} (P_{T,\textrm{topk}}(\cdot|\mathbf{x}) || P_{S,\textrm{topk}}(\cdot|\mathbf{x})) \\
I(\mathbf{x}) &= \mathbf{1}[\textrm{argmax}_y P_T(y|\mathbf{x}) = \textrm{label}(\mathbf{x})] \\
P_{\cdot, \textrm{topk}}(y|\mathbf{x}) &= \frac{\mathbf{1}_{y \in Y_{\textrm{topk}}} P_{\cdot}(y|\mathbf{x})}{\sum_{y \in Y_{\textrm{topk}}} P_{\cdot}(y|\mathbf{x})}; 
Y_{\textrm{topk}} = \textrm{argtopk}_y P_T(y|\mathbf{x})
\end{aligned}
\end{equation}
Here $P_T$ and $P_S$ denote teacher and student label probabilities; $I(\cdot)$ filters out images misaligned with language labels; and $k$ controls the number of most-similar language features for each image. In our implementations, we find a larger $k$ beneficial for OOD generalization, and we choose $k=256$.

We demonstrate the effectiveness of $\mathcal{L}_{\textrm{vlprox}}$ in Tab.~\ref{tab:main_imonly}. We find that by combining $\mathcal{L}_{\textrm{vlprox}}$ with $\mathcal{L}_{\textrm{cls}}$ and $\mathcal{L}_{\textrm{im-cst}}$, we further improve student's ability to generalize towards OOD concepts. Interestingly, we also observe that while $\mathcal{L}_{\textrm{cls}}$ and $\mathcal{L}_{\textrm{vlprox}}$ both explicitly perform V-L alignments, adding them together yields significantly better student performance on $\mathcal{X}_{\textrm{id}}$ and 5-shot $\mathcal{X}_{\textrm{ood}}$ than solely keeping $\mathcal{L}_{\textrm{vlprox}}$. This observation is distinct from those in the traditional model distillation literature~\cite{distillhinton}, where distilling teacher logits alone from vision-only models typically produces good student performance.

\noindent\textbf{Quantifying teacher-student vision-language alignment coherency.} To better understand the vision-language alignment proximity between teacher and student, we propose a metric $\mathcal{M}_{\textrm{vlalign}}$ to quantify such proximity:
\begin{equation}
\begin{aligned}
    \mathcal{M}_{\textrm{vlalign}}(\mathcal{X}, k) &= \frac{\sum_{i=1}^{|\mathcal{X}|} \textrm{\#reverse\_pairs}(\textrm{arrS}(i, k))}{|\mathcal{X}|} \\
    \textrm{arrS}(i,k) &= [||S(\mathbf{x}_i) - T_{\textrm{txt}}(l(y_j))||_2]_{j\in \mathcal{I}(i,k)} \\
    \mathcal{I}(i, k) &= \textrm{argtopk}([-||T_{\textrm{im}}(\mathbf{x}_i) - T_{\textrm{txt}}(l(y_j))||_2)]_{j=1}^{|\mathcal{Y}|})
\end{aligned}
\end{equation}
Here $\textrm{\#reverse\_pairs}(\cdot)$ measures the number of reverse pairs $\{(i,j): a_i > a_j \}$ in an array. Overall, $\mathcal{M}_{\textrm{vlalign}}$ measures the extent to which the distance ordering of the $k$ nearest language features to the teacher image feature differs from the distance ordering of the same $k$ language features with respect to the student image feature. In other words, a lower value of $\mathcal{M}_{\textrm{vlalign}}$ indicates a higher degree of proximity to the teacher's vision-language alignment, as teacher's relative language orderings with respect to each image becomes better preserved.

We assess $\mathcal{M}_{\textrm{vlalign}}$ on different students and present the results in Tab.~\ref{tab:metric_levenshtein}. We find that $\mathcal{L}_{\textrm{im-cst}}$ not only fosters strong student-teacher visual space alignments, but also \textit{implicitly} facilitates effective V-L alignments along this process. This observation illustrates that a better relative and local visual space coherence with the teacher (indicated by better $\mathcal{M}_{\textrm{rel}}$ and $\mathcal{M}_{\textrm{neigh}}$) can lead to enhanced alignment coherence in the V-L space. Additionally, the inclusion of $\mathcal{L}_{\textrm{vlprox}}$ further strengthens teacher-student V-L alignments. Such improved alignments are then effectively transferred to unseen concepts, demonstrating that the student has acquired better V-L alignment structures that generalize well to OOD scenarios.

\begin{table}[t]
\centering
\scriptsize
\setlength{\tabcolsep}{3.0pt}
\begin{tabular}{l|ccc}
    \toprule
    $\mathcal{M}_{\textrm{vlalign}} \downarrow$ & $k=2$ & $k=3$ & $k=5$ \\ \midrule
    {$\mathcal{L}_{\textrm{cls}} + \mathcal{L}_{\textrm{mse}}$} & 0.20~/~0.50 & 0.68~/~1.45 & 2.67~/~4.73\\
    {$\mathcal{L}_{\textrm{cls}}+ \mathcal{L}_{\textrm{mse}}+ \mathcal{L}_{\textrm{im-cst}}$} & 0.18~/~0.43 & 0.62~/~1.3 & 2.52~/~4.24\\
    {$\mathcal{L}_{\textrm{cls}}+ \mathcal{L}_{\textrm{mse}}+ \mathcal{L}_{\textrm{im-cst}} + \mathcal{L}_{\textrm{vlprox}}$} & 0.17~/~0.39 & 0.59~/~1.17 & 2.17~/~4.20 \\
    \bottomrule
\end{tabular}
\caption{We evaluate $\mathcal{M}_{\textrm{vlalign}}$ (lower the better) on different students to measure their proximity with the teacher vision-language alignment structure. $x_1/x_2$ in each entry denote $\mathcal{M}_{\textrm{vlalign}}(\mathcal{X}_{\textrm{train}})$ and $\mathcal{M}_{\textrm{vlalign}}(\mathcal{X}_{\textrm{ood}})$, respectively. Metrics are evaluated on Flower102. More results are in Appendix.}
\label{tab:metric_levenshtein}
\end{table}

\section{Language Representation Enrichment}
\label{sec:img_and_text}

\begin{table*}[t]
\centering
\small
\scriptsize
\setlength{\tabcolsep}{4.8pt}
\begin{tabular}{l|cccccc|c} 
    \toprule
    & CaltechBirds & StanfordCars & Flower102 & Food101 & SUN397 & tiered-ImageNet & \textbf{Average}\\ \midrule
    {$\text{Tab.~\ref{tab:main_imonly} best (no lang enrichment)}$} & 62.3~/~21.6~/~39.0 & 63.9~/~19.8~/~38.5 & 82.7~/~14.6~/~52.0 & 74.3~/~32.0~/~43.2 & \textbf{61.7}~/~21.5~/~34.7 & 67.5~/~20.5~/~35.3 & 68.7~/~21.7~/~40.5\\
    Semantic Details & 62.0~/~\textbf{23.2}~/~40.4 & 63.6~/~20.0~/~37.5 & 82.4~/~17.6~/~52.7 & 74.8~/~33.9~/~43.7 & 60.8~/~23.3~/~36.8 & \textbf{69.8}~/~23.5~/~36.2 & 68.9~/~23.6~/~41.2\\
    Auxiliary Captions & 62.5~/~21.4~/~\textbf{41.0} & 65.5~/~19.0~/~38.1 & 81.9~/~14.3~/~52.5 & 75.4~/~33.3~/~44.0 & 61.6~/~22.1~/~36.9 & 68.7~/~21.0~/~34.4 & 69.3~/~21.9~/~41.2\\
    Prompt Learning & 62.7~/~9.2~/~37.4 & 64.5~/~14.9~/~39.5 & 83.0~/~13.3~/~52.7 & \textbf{75.8}~/~29.6~/~43.6 & 59.3~/~16.6~/~37.5 & 68.5~/~19.4~/~37.5 & 69.0~/~17.2~/~41.4 \\
    Semantics + Caption  & 62.0~/~22.7~/~39.8 & 64.9~/~\textbf{20.4}~/~39.7 & \textbf{83.7}~/~\textbf{18.2}~/~\textbf{53.4} & 75.6~/~\textbf{35.7}~/~42.9 & 61.0~/~\textbf{24.0}~/~37.5 & 68.9~/~\textbf{23.6}~/~35.8 & \textbf{69.4}~/~\textbf{24.1}~/~41.5\\
    Semantics + Prompt & \textbf{63.9}~/~16.7~/~40.8 & \textbf{67.0}~/~19.0~/~\textbf{42.8} & 82.3~/~14.8~/~52.8 & 74.4~/~28.4~/~42.3 & 59.4~/~19.2~/~38.4 & 68.6~/~20.9~/~38.0 & 69.3~/~19.8~/~\textbf{42.5}\\
    Semantics + Caption + Prompt & 62.9~/~16.9~/~37.9 & 65.0~/~16.5~/~40.1 & 82.1~/~13.8~/~52.8 & 75.3~/~31.2~/~\textbf{44.3} & 59.5~/~19.7~/~\textbf{38.9} & 68.1~/~22.7~/~\textbf{38.4} & 68.8~/~20.1~/~42.1\\
    \bottomrule
\end{tabular}
\vspace{-0.5em}
\caption{Comparison between different language representation enrichment strategies. The three numbers $x_1/x_2/x_3$ in each entry denote the evaluation performance
on $\mathcal{X}_{\textrm{id}}$, zero-shot performance on $\mathcal{X}_{\textrm{ood}}$, and 5-shot performance on $\mathcal{X}_{\textrm{ood}}$, respectively.}
\label{tab:main_im_with_text}
\end{table*}

In Sec.~\ref{sec:only_image}, we focused on improving the imitation of teacher's visual space and promoting better coherence with teacher's vision-language alignment. Throughout this process, we kept the language representations fixed. However, the quality of language representations also plays a pivotal role in student learning and inference. Ideally, language representations should be capture precise, finegrained, and meaningful semantic attributes, such that the student can effectively distinguish between different labels. We therefore ask the following question: can we leverage better and richer teacher language representations to further enhance student's OOD generalization ability? We propose the following candidate strategies:

\noindent\textbf{Enriching semantic details of label descriptions by prompting LLMs}. Previously, when we generate language representations $l(y)=\textrm{prompt} + \textrm{description}(y)$ for student training, we adopted a simple strategy. In particular, for the description of a label $y$, we merely used its label name, e.g., ``lotus''. However, these simplistic descriptions overlook many finegrained properties of semantic categories, such as the shape, color, and texture of flowers, along with the description of their petals, leaves, and stems. In addition, we hope to automatically and efficiently generate enriched language descriptions for a wide range of labels, ensuring scalability for an arbitrary number of labels. Motivated by the recent progress on instruction-finetuned large language models (LLMs)~\cite{instructgpt2022,chung2022flant5,sanh2021t0,wei2021finetuned}, which have demonstrated impressive sequence generation abilities given user prompts, we find these models well-suited for our goal. Therefore, we propose to use ChatGPT~\cite{gpt3_2020,instructgpt2022} to generate category descriptions. We prompt ChatGPT with the following instruction: \texttt{Use a single sentence to describe the appearance and shape of \{cls\}. Only describe the shape and appearance}. This allows ChatGPT to generate informative, finegrained, and meaningful descriptions for target classes (e.g., \texttt{large, round, flat leaves; tall, slender stems; delicate petals in shades of pink, white, or yellow}), while keeping sequence lengths within the 77 token limit of the CLIP text encoder. We then set $\textrm{description}(y)$ by concatenating ``a photo of \{cls\}'' with ChatGPT-generated class descriptions. Note that we still keep the same vision-language alignment losses ($\mathcal{L}_{\textrm{cls}}$ and $\mathcal{L}_{\textrm{vlprox}}$) as before.

\noindent\textbf{Augmenting text through auxiliary captions}. Currently, during student training, there is only one language description per category, i.e., $|\{l(y) : (\mathbf{x},y) \in  \mathcal{X}_{\mathrm{train}}  \}| = |\mathcal{Y}_{\textrm{id}}|$. On the other hand, the number of training images significantly exceeds the number of labels, i.e., $|\mathcal{X}_{\textrm{train}}| \gg |\mathcal{Y}_{\textrm{id}}|$. We therefore wish to generate language descriptions for each individual image, such that we can substantially enrich the number of language features during student training, which potentially benefits student performance. To achieve this, we propose using OFA~\cite{ofa2022} to generate captions for each image, resulting in a new dataset $\{(\mathbf{x}, \textrm{cap}(\mathbf{x}), y) :  (\mathbf{x}, y) \in \mathcal{X}_{\textrm{train}}\}$ augmented with captions. During student training, besides using the same vision-language alignment losses $\mathcal{L}_{\textrm{cls}}$ and $\mathcal{L}_{\textrm{vlprox}}$ as before, we also adopt the following auxiliary loss:
\begin{equation}
    \mathcal{L}_{\textrm{cap}}(\mathbf{x}) = \frac{\exp(\cos(S(\mathbf{x}), T_{\textrm{txt}}(\textrm{cap}(\mathbf{x})))/\tau)}{\sum_{(\mathbf{x}',y'):y' \ne y} \exp(\cos(S(\mathbf{x}), T_{\textrm{txt}}(\textrm{cap}(\mathbf{x}')) / \tau)}
\end{equation}
The loss pushes $\mathbf{x}$ and its corresponding caption $\textrm{cap}(\mathbf{x})$ together while pulling away from captions belonging to different categories. Our preliminary experiments show that distinguishing captions belonging to the same category could degrade student performance as they are usually similar. \textit{Note that we only incorporate captions for the auxiliary loss during student training. For student inference and label predictions, we continue to use the same $l(y)$ as before.}

\noindent\textbf{Learning prompts to modify language representations}. In previous experiments, we adopted a simple prompt ``a photo of'' to generate $l(y)$ for student learning. By modifying prompts, we can alter the teacher's language representation space and thereby influence vision-language alignment. Furthermore, in real-world scenarios, student networks often need to continuously learn new concepts and semantics, leading to ongoing updates in their backbones. We therefore explore the potential of performing prompt learning during student training on $\mathcal{Y}_{\textrm{id}}$ and few-shot learning on $\mathcal{Y}_{\textrm{ood}}$. Note that our setting is different from many prior settings on prompt tuning~\cite{coop2022,cocoop2022,dualcoop2022,gao2021clip}, where there are no OOD concepts, and vision backbones are kept fixed. We adopt CoOp~\cite{coop2022} and optimize 8 tokens as our prompt.

\subsection{Analyzing the Efficacy of Different Strategies}
\label{sec:lang_efficacy}

We adopt the aforementioned language-enriching strategies for student learning, and we present the results in Tab.~\ref{tab:main_im_with_text}. We find that combining LLM-enriched label descriptions with auxiliary captions yields the best OOD generalization. However, upon analyzing their individual effectiveness, we find that \textit{LLM-enriched label descriptions provide significantly better zero-shot OOD benefit than auxiliary captions, and solely relying on auxiliary captions only marginally improves zero-shot OOD generalizability}. Upon further analysis, we find that many generated captions only broadly describe objects and are much less informative than ChatGPT-generated descriptions for distinguishing fine-grained categories. For instance, in the StanfordCars dataset, a generated caption for an \texttt{Acura Integra Type R 2001} image is \texttt{a white car is parked in a field}, and solely relying on the white color provides little information to distinguish different car categories. Consequently, captions have limited impact on enhancing the generalizability of student's vision-language alignment structures.

We also observe that learning prompts during student training harms its zero-shot OOD generalization performance, suggesting that this approach leads to the overfitting of vision-language alignments on the training categories. However, when provided with a small number of samples from OOD categories, the prompts can quickly adapt to these novel concepts, enabling students to achieve the best few-shot OOD performance overall.

\subsection{Comparing Language Representation Spaces Before and After Enrichment}

To gain further insights into the efficacy of our language enrichment strategies, in this section, we analyze the changes in language space structures before and after adopting such strategies. On the out-of-distribution evaluation dataset of Flower102, we obtain text features $\{T_{\textrm{txt}}(l_{\textrm{new}}(y))\}_{y=1}^{|\mathcal{Y_{\textrm{ood}}}|}$ and $\{T_{\textrm{txt}}(l_{\textrm{old}}(y))\}_{y=1}^{|\mathcal{Y_{\textrm{ood}}}|}$, where $l_{\textrm{new}}$ denotes the language generation function with LLM enrichment, and $l_{\textrm{old}}$ uses simple label names as label descriptions. We also obtain the average caption features for each class, namely $\{\textrm{mean}[\{T_{\textrm{txt}}(\textrm{cap}(\mathbf{x})): \textrm{label}(\mathbf{x})=y\}]\}_{y=1}^{|\mathcal{Y_{\textrm{ood}}}|}$. We then mean-center these 3 sets of text features, perform Singular-Value Decomposition, and plot the resulting eigenvalues in Fig.~\ref{fig:text_eigenvalues}. We observe that language features of $l_{\textrm{new}}$ contain many more large eigenvalues than $l_{\textrm{old}}$, demonstrating that the text space generated by $l_{\textrm{new}}$ confers more independent and meaningful attributes to effectively distinguish between different classes. On the other hand, the language space of auxiliary captions only contains a small number of meaningful attributes, indicating that auxiliary captions are not very helpful for distinguishing between different labels. Additionally, in conjunction with Tab.~\ref{tab:main_im_with_text}, we find that the advantages of richer attributes are concealed when evaluating students on the in-distribution classes, since students tend to overfit the existing attributes of these labels during training. However, for OOD scenarios, the presence of more finegrained attributes allows OOD text features to be more precisely aligned with image features, especially when no vision-language alignment training is done on these OOD samples (see Fig.~\ref{fig:visualization_teaser_umap} for an illustration). As a result, students exhibit better OOD generalization ability.

Furthermore, we analyze the average cosine similarity between all pairs of text features within each of the three sets we introduced earlier. We present the results in Tab.~\ref{tab:sem_enrichment_cos_similarity}. We observe that LLM-enriched semantic details allow text features to naturally separate further apart from each other, making it easier to distinguish between different classes. In contrast, auxiliary captions show limited effectiveness in achieving such separation between classes.

\begin{minipage}{\textwidth}
\hspace{-1.5em}
\begin{minipage}[t]{.23\textwidth}
\includegraphics[width=\textwidth]{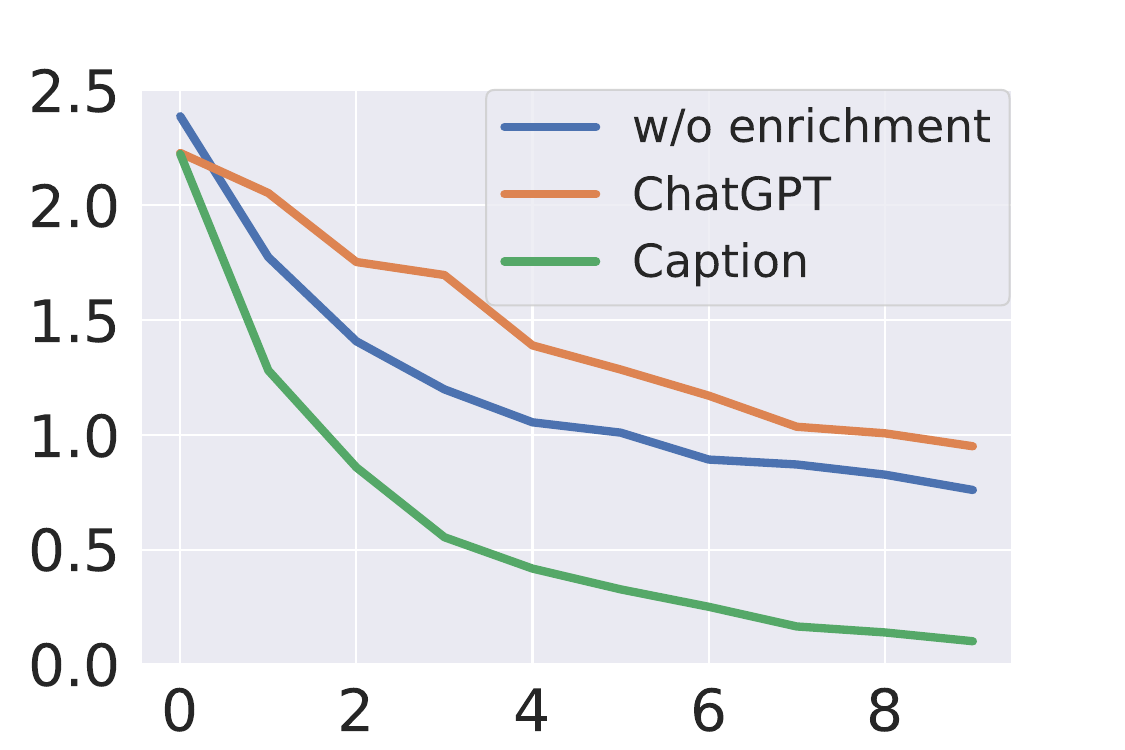}
\vspace{-1em}
\captionof{figure}{Top 10 eigenvalues of text features.}
\label{fig:text_eigenvalues}
\end{minipage}
\begin{minipage}[t]{.01\textwidth}
\ 
\end{minipage}
\begin{minipage}[t]{.23\textwidth}
\vspace{-6.5em}
\centering
\small
\setlength{\tabcolsep}{1.5pt}
\begin{tabular}{c|cc}
    \toprule
     & Cos. Sim.\\ \midrule
    w/o enrichment & 0.5156 \\
    ChatGPT & 0.4462 \\
    Caption & 0.5572\\
    \bottomrule
\end{tabular}
\captionof{table}{Average cosine similarity between all pairs of text features.}
\label{tab:sem_enrichment_cos_similarity}
\end{minipage}
\end{minipage}

\section{Ablations}
\label{sec:ablations}
In this section, we present further ablation studies to complement our findings in Sec.~\ref{sec:only_image} and Sec.~\ref{sec:img_and_text}.

\noindent\textbf{Control semantic details by prompting LLMs}. In Sec.~\ref{sec:img_and_text}, we found that descriptive semantic details in $l(y)$ are particularly helpful for student generalization on out-of-distribution concepts. We thus ask the following question: how specific should semantic details be, and which semantic details are helpful? We construct the following prompts to \textit{control how ChatGPT generates semantic details} (example generations in Appendix):
\textbf{More Succinct}: ``Use a single sentence to broadly describe the appearance and shape of \{cls\}. Don't give too much details. Only describe the shape and appearance.''
\textbf{More Detailed}: ``Use a single sentence and short, simple, descriptive phrases to describe the detailed appearance and detailed shape of \{cls\}.'' 
\textbf{More Distinct}: ``Use a single sentence to describe the unique, distinctive appearance and shape of \{cls\}. Only describe the unique, distinctive shape and appearance.''

We compare these prompts in Tab.~\ref{tab:ablation_chatgpt}. Interestingly, we observe that generating more detailed semantic descriptions on labels does not always perform better. We conjecture that this is because (1) LLM-generated details are not grounded in specific images, causing some attributes to be invisible and confusing the students; (2) the teacher CLIP is trained on LAION~\cite{schuhmann2021laion}, where most language descriptions do not contain many finegrained appearance details, so CLIP's text embeddings are not very sensitive to some of these details. Additionally, we find that explicitly prompting ChatGPT to generate more concise text descriptions could be still helpful. Upon further analysis, we find that the resulting generations remain highly descriptive, albeit with slightly fewer details (e.g., when describing a \texttt{trumbone}, the more concise description becomes \texttt{a brass instrument with a long cylindrical tube curved into an elongated S shape with a flared bell at the end}, whereas under our original prompt, additional details like \texttt{a sliding U-shaped section called the slide} are included).

\begin{table}[t]
\centering
\scriptsize
\setlength{\tabcolsep}{6.0pt}
\begin{tabular}{l|cc}
    \toprule
     & StanfordCars & tiered-ImageNet \\ \midrule
     No language enrichment & \textbf{63.9}~/~19.8~/~\textbf{38.5} & 67.5~/~20.5~/~35.3 \\
    Prompt in Sec.~\ref{sec:img_and_text} & 63.3~/~\textbf{20.0}~/~37.5 & \textbf{69.8}~/~23.5~/~36.2 \\
    More Succinct & 63.0~/~18.9~/~37.6 & 68.8~/~23.1~/~36.8 \\
    More Detailed & 62.9~/~19.0~/~35.1 & 69.3~/~\textbf{24.2}~/~\textbf{37.7}\\
    More Distinct & \textbf{63.9}~/~19.7~/~37.1 & 69.2~/~23.3~/~37.0 \\
    \bottomrule
\end{tabular}
\caption{Results on leveraging different prompts to control semantic details of label descriptions generated by ChatGPT.}
\label{tab:ablation_chatgpt}
\end{table}

\begin{table}[t]
\centering
\scriptsize
\setlength{\tabcolsep}{1.4pt}
\begin{tabular}{cccc|ccc}
    \toprule
    $\mathcal{L}_{\textrm{cls}}$ & $\mathcal{L}_{\textrm{im-cst}}$ & $\mathcal{L}_{\textrm{vlprox}}$ & Semantics & Flower102 & CaltechBirds & SUN397\\ \midrule
    $\checkmark$ & & & & 77.9~/~5.3~/~31.3 & 19.7~/~6.2~/~10.6 & 39.6~/~8.6~/~13.5 \\
    $\checkmark$ & $\checkmark$ & & & 78.1~/~11.0~/~46.0 & 21.5~/~7.7~/~11.1 & 44.2~/~13.7~/~18.3 \\
    $\checkmark$ & $\checkmark$ &  & $\checkmark$ & 77.8~/~11.9~/~46.5 & \textbf{22.2}~/~\textbf{8.8}~/~12.7 & 44.3~/~14.7~/~19.5 \\
    $\checkmark$ & $\checkmark$ & $\checkmark$ & $\checkmark$ & \textbf{78.9}~/~\textbf{13.1}~/~\textbf{47.7} & \textbf{22.2}~/~8.5~/~\textbf{13.1} & \textbf{45.0}~/~\textbf{15.4}~/~\textbf{21.0} \\
    \bottomrule
\end{tabular}
\caption{Distilling CLIP ViT-L/14 into a student ViT-B/32 network. In each entry, $x_1/x_2/x_3$ denote $\mathcal{X}_{\textrm{id}}$ / zero-shot $\mathcal{X}_{\textrm{ood}}$ / 5-shot $\mathcal{X}_{\textrm{ood}}$ performance.}
\label{tab:ablation_vit}
\end{table}

\begin{table}[t]
\centering
\scriptsize
\setlength{\tabcolsep}{5.0pt}
\begin{tabular}{l|cc}
    \toprule
     & Flower102 & tiered-ImageNet \\ \midrule
     Tab.~\ref{tab:main_im_with_text} best & 82.4~/~17.6~/~52.7 & 69.8~/~23.5~/~36.2 \\
    No filtering in $\mathcal{L}_{\textrm{vlprox}}$ & 83.3~/~16.8~/~51.5 & 69.6~/~23.2~/~35.3 \\
    $k=10$ for $\mathcal{L}_{\textrm{vlprox}}$ & 83.4~/~17.0~/~51.4 & 68.4~/~22.6~/~34.6 \\
    $k=1$ for $\mathcal{L}_{\textrm{vlprox}}$ & 81.4~/~15.9~/~51.4 & 66.7~/~21.0~/~35.4 \\
    With filtering in $\mathcal{L}_{\textrm{im-cst}}$ & 83.6~/~16.6~/~51.0 & 68.5~/~23.1~/~33.4 \\
    \bottomrule
\end{tabular}
\caption{Ablation studies on (1) different designs of $\mathcal{L}_{\textrm{vlprox}}$, and (2) whether to imitate the entirety of the teacher's visual space in $\mathcal{L}_{\textrm{im-cst}}$.}
\label{tab:ablation_proximal_img_txt}
\end{table}

\noindent\textbf{Different designs on visual space and vision-language alignments}. We ablate on our designs of visual space and vision-language alignments and present the results in Tab.~\ref{tab:ablation_proximal_img_txt}. We first explore various designs of $\mathcal{L}_{\textrm{vlprox}}$, which aims to improve the \textit{vision-language} alignment coherency with the teacher. We find that enhancing coherency with a greater number of most-similar language features by increasing $k$ improves student OOD generalization. Additionally, it is also beneficial to filter out teacher's image features that are misaligned with their corresponding language labels. Interestingly, we observe a different pattern when considering the teacher-student \textit{visual space} alignments. In this case, we did not find filtering out misaligned image features helpful, and imitating the entirety of teacher's visual space yields better student OOD generalization.

\noindent\textbf{Different student visual backbones}. In our prior experiments, we adopted ResNet as our student visual backbone. In this section, we further investigate whether our findings are consistent across different student network architectures. We adopt ViT-B/32~\cite{vit2021} as our student, while keeping CLIP ViT-L/14 as our VLM teacher. Results are shown in Tab.~\ref{tab:ablation_vit}. We observe that even though we apply strong data augmentations (RandAugment) when training the student ViT-B/32, it still suffers from severe overfitting (the training accuracy on $\mathcal{X}_{\textrm{train}}$ is $>90\%$, which is significantly higher than the accuracy on $\mathcal{X}_{\textrm{id}}$ in Tab.~\ref{tab:ablation_vit}). The in-distribution and OOD generalization performance of the ViT-B/32 student are also worse than the ResNet student. Nevertheless, we still observe that better teacher-student visual space alignment, improved coherence in vision-language alignment, and language representation enrichment all enhance student's OOD generalization ability, aligning with our previous findings.

\begin{table}[t]
\centering
\small
\setlength{\tabcolsep}{4.0pt}
\begin{tabular}{l|ccc}
    \toprule
    & CaltechBirds & SUN397 & tiered-ImageNet \\ \midrule
    5-shot Finetune & \textbf{40.36} & \textbf{36.81} & \textbf{36.17} \\
    5-shot Retrieval & 24.39 & 24.40 & 24.27 \\
    \bottomrule
\end{tabular}
\caption{Comparison between finetuning student visual backbone vs. training-free retrieval on $\mathcal{X}_{\textrm{ood}}$. Finetuning student backbones significantly outperforms training-free retrieval. }
\label{tab:ablation_finetuneretrieval}
\end{table}

\begin{table}[t]
\centering
\small
\begin{tabular}{l|ccc}
    \toprule
    & $\mathcal{X}_{\textrm{ood1}}$  & $\mathcal{X}_{\textrm{ood2}}$ \\ \midrule
    0-shot & 25.11 & 18.01 \\
    5-shot on $\mathcal{X}_{\textrm{ood1}}$ & \textbf{62.79} & \textbf{18.24}\\
    \bottomrule
\end{tabular}
\caption{Comparison between the student model's zero-shot generalization performance on $\mathcal{X}_{\textrm{ood2}}$ before and after few-shot finetuning on $\mathcal{X}_{\textrm{ood1}}$. }
\label{tab:ablation_continuous_adaptation}
\end{table}

\noindent\textbf{Few-shot learning strategies on $\mathcal{X}_{\textrm{ood}}$}. Previously, we finetune the student visual networks during few shot learning on $\mathcal{X}_{\textrm{ood}}$. Alternatively, we could adopt a finetune-free retrieval-based strategy like \cite{tipadapter2022}, which has been shown effective on large VLMs. We compare these two strategies in Tab.~\ref{tab:ablation_finetuneretrieval}. Further implementation details are presented in Appendix. Interestingly, we observe that the retrieval-based strategy significantly underperforms finetuning student visual backbones. We hypothesize that this phenomenon arises because our student networks are trained on small to medium-scale datasets, and have been exposed to far fewer concepts than their teacher VLMs. Consequently, student's image feature structure and vision-language alignment structure are less generalizable on OOD concepts, making them less suitable for retrieval-based approaches.

\noindent\textbf{Zero-shot OOD generalization ability after few-shot learning}. In the real world, it is essential for student networks to continuously adapt to new concepts, and we aim to find a student learning strategy that accomplishes such goal. Since we finetune student visual backbones during few-shot learning on $\mathcal{X}_{\textrm{ood}}$, we would like to know whether finetuned student backbones overfit seen concepts and exhibit weaker zero-shot generalization ability when encountering novel unseen concepts again. We conduct an experiment on Flower102, where we split $\mathcal{Y}_{\textrm{ood}}$ into two equal sets, and then split $\mathcal{X}_{\textrm{ood}}$ into $\mathcal{X}_{\textrm{ood1}}$ and $\mathcal{X}_{\textrm{ood2}}$ accordingly. We then select the best student model and evaluate it on $\mathcal{X}_{\textrm{ood2}}$ both before and after few-shot finetuning it on $\mathcal{X}_{\textrm{ood1}}$. Results are presented in Tab.~\ref{tab:ablation_continuous_adaptation}. We observe that the student's zero-shot OOD generalization ability slightly improves after few-shot learning, demonstrating that students can continuously adapt to novel concepts.

\section{Application}
\label{sec:applications}

\begin{table}
\centering
\small
\begin{subtable}{0.45\textwidth}
\centering
\setlength{\tabcolsep}{3.5pt}
\begin{tabular}{l|ccc}
    \toprule
     & $\mathcal{X}_{\textrm{id}}$  & $\mathcal{Y}_{\textrm{id}}$ on $\mathcal{X}_{\textrm{ood}}$ & $\mathcal{Y}_{\textrm{ood}}$ on $\mathcal{X}_{\textrm{ood}}$\\ \midrule
    Closed-Set & 96.4~/~96.5 & NA~/~86.2 & NA~/~87.7 \\
    $\mathcal{L}_{\textrm{cls}}$ & 96.9~/~97.2 & 79.3~/~85.3 & 71.7~/~87.3 \\
    + $\mathcal{L}_{\textrm{im-cst}}$ & \textbf{99.2}~/~\textbf{99.2} & 84.0~/~91.9 & 76.3~/~88.3 \\
    + Semantic Enrich & 98.2~/~99.0 & \textbf{84.3}~/~\textbf{92.0} & \textbf{83.0}~/~\textbf{89.6} \\
    \bottomrule
\end{tabular}
\caption{Overall accuracy over all YCB objects}
\vspace{0.4em}
\end{subtable}
\begin{subtable}{0.45\textwidth}
\centering
\setlength{\tabcolsep}{3.5pt}
\begin{tabular}{l|ccc}
    \toprule
     & $\mathcal{X}_{\textrm{id}}$  & $\mathcal{Y}_{\textrm{id}}$ on $\mathcal{X}_{\textrm{ood}}$ & $\mathcal{Y}_{\textrm{ood}}$ on $\mathcal{X}_{\textrm{ood}}$\\ \midrule
    Closed-Set & 91.6~/~91.9 & NA~/~57.8 & NA~/~35.9 \\
    $\mathcal{L}_{\textrm{cls}}$ & 94.0~/~94.4 & 46.5~/~54.7 & 23.3~/~32.8 \\
    + $\mathcal{L}_{\textrm{im-cst}}$ & \textbf{98.1}~/~\textbf{98.0} & 55.3~/~\textbf{70.8} & \textbf{23.7}~/~47.3 \\
    + Semantic Enrich & 97.2~/~97.4 & \textbf{55.6}~/~\textbf{70.8} & 11.7~/~\textbf{50.7} \\
    \bottomrule
\end{tabular}
\caption{F1-measure over objects that exist in observations.}
\end{subtable}
\caption{Results on grasp feasibility prediction in the PickClutter task. In each entry, $x_1/x_2$ denote zero-shot and few-shot evaluation results, respectively. ($\mathcal{L}_{\textrm{vlprox}}$ is not available for this experiment since multiple objects exists in an observation)}
\label{tab:robotics_results}
\end{table}

\begin{wrapfigure}{r}{0.23\textwidth}
    \centering
\includegraphics[width=0.23\textwidth]{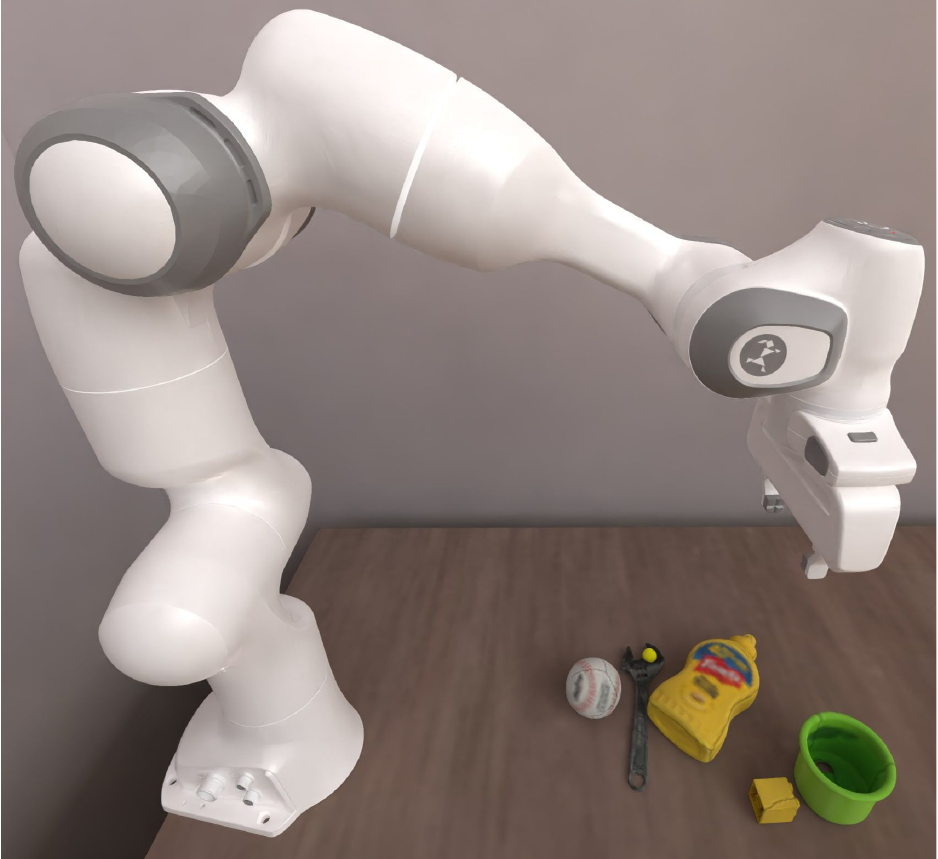}
    \vspace{-1em}
    \caption{Illustration of the PickClutter task. Given the name of a target object, the robot agent needs to decide whether the corresponding object is graspable based on the current visual observation.} %
    \label{fig:robotics_illustration}
\end{wrapfigure}

In this section, we demonstrate that we can adopt our previous findings for improving student's OOD generalization ability towards novel tasks and domains. We augment the PickClutter task from a robot object manipulation skill benchmark ManiSkill2~\cite{gu2023maniskill2} with language, where given the name of a YCB object~\cite{calli2015ycb}, a robot needs to detect whether it exists among a pile of objects given the current visual observation captured from a hand camera, and if exists, picks up this object. The task is illustrated in Fig.~\ref{fig:robotics_illustration}. We randomly sample different configurations of objects, and given observations in each configuration, the student network outputs whether it is feasible to grasp each YCB object. This setup resembles observation-based affordance prediction in works such as SayCan~\cite{ahn2022can}.
We select 26 visually-distinctive YCB objects and split them equally into $\mathcal{Y}_{\textrm{id}}$ and $\mathcal{Y}_{\textrm{ood}}$. We then construct the datasets such that $\mathcal{X}_{\textrm{train}}$ and $\mathcal{X}_{\textrm{id}}$ only contain objects in $\mathcal{Y}_{\textrm{id}}$, while $\mathcal{X}_{\textrm{ood}}$ contains objects in both $\mathcal{Y}_{\textrm{id}}$ and $\mathcal{Y}_{\textrm{ood}}$. 
We use 3000 scenes for training and 50 scenes for few shot learning. We adopt EfficientNet~\cite{tan2019efficientnet} as the student network and CLIP ViT-L-14 as the teacher. As we are in a multi-label classification setting, our preliminary experiments show that having positive and negative prompts like \cite{dualcoop2022} and learning these prompts during student training can significantly improve student performance, so we adopt these techniques in our experiments. We calculate two metrics: (1) overall accuracy across all YCB objects in the label set (i.e., $\mathcal{Y}_{\textrm{id}}$ for $\mathcal{X}_{\textrm{id}}$, and $\mathcal{Y}_{\textrm{id}} \cup \mathcal{Y}_{\textrm{ood}}$ for $\mathcal{X}_{\textrm{ood}}$); (2) F1-measure calculated over the objects present in an observation, averaged over all observations. Further details are presented in Appendix.
\setlength{\columnsep}{12pt}

We present the results in Tab.~\ref{tab:robotics_results}. Consistent with our previous findings, we find that improving teacher-student visual space alignments through $\mathcal{L}_{\textrm{im-cst}}$ significantly enhances the student's generalization ability on OOD objects. Additionally, leveraging language models to enrich the semantic details of object descriptions benefits few-shot OOD generalization. Interestingly, for zero-shot OOD generalization, while this enrichment improves the overall prediction accuracy for novel objects, it adversely affects the recall (and thus the F1-measure) on these objects. Further analysis reveals that students tend to ignore objects unseen during training, resulting in lower recall. However, with just a few examples of novel objects, students achieve significantly better recall on these objects.

\section{Conclusion}

In this work, we studied distillation of large teacher vision-language models into lightweight student models by focusing on open-vocabulary out-of-distribution (OOD) generalization for object classification using small to medium-scale datasets. We investigated strengthening students' OOD generalization ability from two key perspectives: first, by better imitating teacher's visual representation space and carefully promoting better teacher-student vision-language alignment coherence; and second, by enhancing the teacher's language representations with informative and meaningful semantic attributes to effectively differentiate between different labels. We analyzed the efficacy and impact of our techniques by introducing metrics and conducting a comprehensive experimental analysis. Along this process, we significantly improve student's zero-shot and few-shot generalization performance on open-vocabulary OOD classification tasks.

\section*{Acknowledgements}

We sincerely thank Jiayuan Gu from UC San Diego for their valuable discussions and feedback. This work is in part supported by Qualcomm AI and AI Institute for Learning-Enabled Optimization at Scale (TI-LOS).

{\small
\bibliographystyle{ieee_fullname}
\bibliography{egbib}
}

\setcounter{section}{0}
\renewcommand{\thesection}{\Alph{section}}
\newpage
\section*{Appendix}

\section{Dataset Statistics}
\label{sec:app_dataset_stats}

\begin{table}[h]
\centering
\scriptsize
\setlength{\tabcolsep}{1.3pt}
\begin{tabular}{l|cccccc} 
    \toprule
    & CaltechBirds & StanfordCars & Flower102 & Food101 & SUN397 & tiered-ImageNet \\ \midrule
    $|\mathcal{X}_{\textrm{train}}|$ & 4122 & 2874 & 3112 & 35700 & 38663 & 314108 \\
    $|\mathcal{X}_{\textrm{id}}|$ & 1740 & 1164 & 1303 & 15300 & 16444 & 134587 \\
    $|\mathcal{X}_{\textrm{ood}}|$ & 5926 & 4106 & 3774 & 50000 & 53647 & 124261 \\
    $|\mathcal{Y}_{\textrm{id}}|$ & 100 & 98 & 51 & 51 & 200 & 351 \\
    $|\mathcal{Y}_{\textrm{ood}}|$ & 100 & 98 & 51 & 50 & 197 & 97 \\
    \bottomrule
\end{tabular}
\caption{Dataset Statistics for our main experiments. Terminologies follow Sec.~\ref{sec:overview}.}
\label{tab:dataset_stats}
\end{table}

In Tab.~\ref{tab:dataset_stats}, we provide dataset split statistics for the experiments introduced in Sec.~\ref{sec:overview} of the main paper.

\section{Training Details and Hyperparameters for Main Experiments}
\label{sec:app_hyperparams}

During agent training on $\mathcal{X}_{\textrm{train}}$, for CaltechBirds, Stanford Cars, and Flower102, due to their relatively small dataset size, we train agents for 450 epochs to ensure convergence; for Food101, SUN397, and tiered-ImageNet, we train agents for 90 epochs. For ResNet-based students, we adopt an initial learning rate of 0.05 with batch size 128, which is decreased to $0.005$ after $1/3$ of training epochs and to $0.0005$ after $2/3$ of training epochs. We adopt standard data augmentation (Random cropping image to 224x224, and random horizontal flip). For ViT-based students, we adopt a one-cycle learning rate schedule with a learning-rate peak of $0.0002$. We also apply RandAugment~\cite{cubuk2020randaugment}, a strong data augmentation method, to cope with overfitting. However, even with RandAugment, we still observe that the performance on $\mathcal{X}_{\textrm{ood}}$ starts to decrease at some point in training, suggesting that it is challenging to train ViT students on small to medium-scale datasets to obtain good out-of-distribution generalizability.

During few-shot learning on $\mathcal{X}_{\textrm{ood}}$, we adopt a balanced training batch, where at most half of samples come from few-shot samples on $\mathcal{X}_{\textrm{ood}}$ and the rest of samples come from $\mathcal{X}_{\textrm{train}}$. For CaltechBirds, Stanford Cars, and Flower102, we train agents for 100 epochs. For Food101, SUN397, and tiered-ImageNet, we train agents for 20 epochs. We adopt a one-cycle learning rate schedule for all student networks. For ResNet-based students, the learning-rate peak is $0.003$. For ViT-based students, the learning-rate peak is $0.0001$. We adopt the same data augment strategy as before.

All experiment results are obtained from the average performance of the last 5 epochs of training.

\section{More Metric Evaluations}
\label{sec:app_more_metrics}

\begin{table*}[ht]
\centering
\scriptsize
\begin{tabular}{ll|cccccc}
\toprule
\multicolumn{2}{l|}{$\mathcal{M}_{\textrm{rel}} \uparrow$} & CaltechBirds & StanfordCars & Flower102 & Food101 & SUN397 & tiered-ImageNet \\ \midrule
\multirow{2}{*}{$\mathcal{L}_{\textrm{cls}}+\mathcal{L}_{\textrm{mse}}$} & $\mathcal{X}_{\textrm{train}}$ & 0.035 & 0.059 & 0.030 & 0.005 & 0.009 & 0.013\\
            & $\mathcal{X}_{\textrm{ood}}$ & 0.014 & 0.017 & 0.004 & 0.002 & 0.003 & 0.007\\ \midrule
\multirow{2}{*}{$\mathcal{L}_{\textrm{cls}}+\mathcal{L}_{\textrm{mse}}+\mathcal{L}_{\textrm{im-cst}}$} & $\mathcal{X}_{\textrm{train}}$ & 0.592 & 0.669 & 0.305 & 0.108 & 0.062 & 0.088\\
             & $\mathcal{X}_{\textrm{ood}}$ & 0.081 & 0.106 & 0.022 & 0.028 & 0.019 & 0.041\\         
\bottomrule
\end{tabular}
\caption{We evaluate $\mathcal{M}_{\textrm{rel}}$ (higher the better) on different datasets to measure how students preserve the relative feature relationships of the teacher's visual representation space.}
\label{tab:app_mrel_results}
\vspace{1em}
\begin{tabular}{ll|cccccc}
\toprule
\multicolumn{2}{l|}{$\mathcal{M}_{\textrm{neigh}} \uparrow$} & CaltechBirds & StanfordCars & Flower102 & Food101 & SUN397 & tiered-ImageNet \\ \midrule
\multirow{3}{*}{$\mathcal{L}_{\textrm{cls}}+\mathcal{L}_{\textrm{mse}}$} & $k=3$ & 0.11~/~0.03 & 0.15~/~0.04 & 0.13~/~0.06 & 0.01~/~0.00 & 0.03~/~0.01 & 0.03~/~0.01\\
                                                        & $k=5$ & 0.15~/~0.04 & 0.21~/~0.05 & 0.18~/~0.07 & 0.02~/~0.01 & 0.04~/~0.01 & 0.05~/~0.01\\ 
                                                        & $k=10$ & 0.24~/~0.06 & 0.30~/~0.07 & 0.27~/~0.08 & 0.03~/~0.01 & 0.06~/~0.02 & 0.07~/~0.02\\ \midrule
\multirow{3}{*}{$\mathcal{L}_{\textrm{cls}}+\mathcal{L}_{\textrm{mse}}+\mathcal{L}_{\textrm{im-cst}}$} & $k=3$ & 0.22~/~0.05 & 0.32~/~0.08 & 0.20~/~0.10 & 0.04~/~0.01 & 0.06~/~0.02 & 0.07~/~0.02\\
             & $k=5$ & 0.28~/~0.06 & 0.36~/~0.09 & 0.25~/~0.11 & 0.05~/~0.01 & 0.07~/~0.02 & 0.09~/~0.03\\
             & $k=10$ & 0.35~/~0.09 & 0.43~/~0.11 & 0.34~/~0.13 & 0.08~/~0.02 & 0.11~/~0.03 & 0.13~/~0.03\\
\bottomrule
\end{tabular}
\caption{We evaluate $\mathcal{M}_{\textrm{neigh}}$ (higher the better) on different datasets to measure how students preserve the local structure of the teacher's visual representation space. $x_1/x_2$ in each entry denote $\mathcal{M}_{\textrm{neigh}}(\mathcal{X}_{\textrm{train}})$ and $\mathcal{M}_{\textrm{neigh}}(\mathcal{X}_{\textrm{ood}})$, respectively.}
\label{tab:app_mneigh_results}
\vspace{1em}
\begin{tabular}{ll|cccccc}
\toprule
\multicolumn{2}{l|}{$\mathcal{M}_{\textrm{vlalign}} \downarrow$} & CaltechBirds & StanfordCars & Flower102 & Food101 & SUN397 & tiered-ImageNet \\ \midrule
\multirow{3}{*}{$\mathcal{L}_{\textrm{cls}}+\mathcal{L}_{\textrm{mse}}$} & $k=2$ & 0.20~/~0.39 & 0.20~/~0.37 & 0.20~/~0.50 & 0.09~/~0.38 & 0.21~/~0.38 & 0.21~/~0.42\\
                                                        & $k=3$ & 0.72~/~1.17 & 0.57~/~1.15 & 0.68~/~1.45 & 0.51~/~1.16 & 0.72~/~1.17 & 0.75~/~1.30\\ 
                                                        & $k=5$ & 2.47~/~3.91 & 2.06~/~3.70 & 2.67~/~4.73 & 2.33~/~3.98 & 2.78~/~3.96 & 2.81~/~4.35\\ \midrule
\multirow{3}{*}{$\mathcal{L}_{\textrm{cls}}+\mathcal{L}_{\textrm{mse}}+\mathcal{L}_{\textrm{im-cst}}$} & $k=2$ & 0.21~/~0.37 & 0.19~/~0.33 & 0.18~/~0.43 & 0.10~/~0.29 & 0.20~/~0.35 & 0.21~/~0.39\\
             & $k=3$ & 0.71~/~1.15 & 0.54~/~1.07 & 0.62~/~1.30 & 0.49~/~0.93 & 0.67~/~1.08 & 0.72~/~1.22\\
             & $k=5$ & 2.30~/~3.74 & 1.89~/~3.53 & 2.52~/~4.24 & 2.19~/~3.42 & 2.55~/~3.68 & 2.72~/~4.11\\ \midrule
\multirow{3}{*}{$\mathcal{L}_{\textrm{cls}}+\mathcal{L}_{\textrm{mse}}+\mathcal{L}_{\textrm{im-cst}}$ + $\mathcal{L}_{\textrm{vlprox}}$} & $k=2$ & 0.19~/~0.35 & 0.17~/~0.37 & 0.17~/~0.39 & 0.09~/~0.29 & 0.19~/~0.34 & 0.19~/~0.36\\
             & $k=3$ & 0.66~/~1.06 & 0.46~/~1.11 & 0.59~/~1.17 & 0.45~/~0.93 & 0.65~/~1.06 & 0.67~/~1.12\\
             & $k=5$ & 2.18~/~3.52 & 1.70~/~3.61 & 2.17~/~4.20 & 2.06~/~3.35 & 2.44~/~3.60 & 2.54~/~3.82\\
\bottomrule
\end{tabular}
\caption{We evaluate $\mathcal{M}_{\textrm{vlalign}}$ (lower the better) on different datasets to measure the proximity between student and teacher vision-language alignment structures. $x_1/x_2$ in each entry denote $\mathcal{M}_{\textrm{vlalign}}(\mathcal{X}_{\textrm{train}})$ and $\mathcal{M}_{\textrm{vlalign}}(\mathcal{X}_{\textrm{ood}})$, respectively.}
\label{tab:app_mvlalign_results}
\end{table*}

In this section, we evaluate $\mathcal{M}_{\textrm{rel}}$, $\mathcal{M}_{\textrm{neigh}}$, and $\mathcal{M}_{\textrm{vlalign}}$ on more datasets to complement our results on teacher-student visual space and vision-language alignment in Sec.~\ref{sec:only_image}. $\mathcal{M}_{\textrm{rel}}$ results are presented in Tab.~\ref{tab:app_mrel_results}. $\mathcal{M}_{\textrm{neigh}}$ results are presented in Tab.~\ref{tab:app_mneigh_results}. $\mathcal{M}_{\textrm{vlalign}}$ results are presented in Tab.~\ref{tab:app_mvlalign_results}.

\section{Example ChatGPT Label Description Generations from Different Prompts}
\label{sec:app_example_chatgpt}

In Sec.~\ref{sec:ablations}, we investigated leveraging different prompts to control the level of semantic details ChatGPT generates for the description of each label, and how these different generated label descriptions impact student performance. In this section, we provide a list of example label descriptions generated using various prompts.

\begin{itemize}
    \item \textbf{Original prompt} in Sec.~\ref{sec:img_and_text}: ``Use a single sentence to describe the appearance and shape of \{cls\}. Only describe the shape and appearance.''
    \begin{itemize}
        \item \texttt{The Egyptian Cat, or Egyptian Mau, is a medium-sized feline with a lithe and muscular body, a short, spotted coat in colors such as silver or bronze, and large, almond-shaped green eyes.}
        \item \texttt{The space bar is a rectangular key on a computer keyboard that is longer horizontally than other keys and has a slightly concave shape for the thumb.}
        \item \texttt{Hair spray typically comes in a cylindrical container with a nozzle on top for spraying, and may vary in size and color depending on the brand and type.}
    \end{itemize}
    \item \textbf{``More Succinct'' prompt} in Sec.~\ref{sec:ablations}: ``Use a single sentence to broadly describe the appearance and shape of \{cls\}. Don’t give too much details. Only describe the shape and appearance.''
    \begin{itemize}
        \item \texttt{The Egyptian cat, also known as the Egyptian Mau, has a slender, athletic body with a spotted coat and large, expressive eyes.}
        \item \texttt{The space bar is a rectangular-shaped key located at the bottom-center of a standard computer keyboard.}
        \item \texttt{Hair spray is typically contained in an aerosol can with a long, narrow nozzle used for spraying the product onto hair.}
    \end{itemize}
    \item \textbf{``More Detailed'' prompt} in Sec.~\ref{sec:ablations}: ``Use a single sentence and short, simple, descriptive phrases to describe the detailed appearance and detailed shape of \{cls\}.''
    \begin{itemize}
        \item \texttt{The Egyptian Cat has a slender, muscular build, a distinctively slender face with large pointy ears, almond-shaped green or gold eyes, and a short, smooth coat in colors such as black, white, or tortoiseshell.}
        \item \texttt{The space bar is a horizontal, elongated key located at the bottom center of a computer keyboard, typically featuring a smooth, concave surface, and comes in various sizes and colors for different models.}
        \item \texttt{The hair spray is a pressurized aerosol can or pump bottle with a plastic or metal container, typically transparent or colored, and comes in various strengths and formulas, used for holding and styling the hair in place.}
    \end{itemize}
    \item \textbf{``More Distinct'' prompt} in Sec.~\ref{sec:ablations}: ``Use a single sentence to describe the unique, distinctive appearance and shape of \{cls\}. Only describe the unique, distinctive shape and appearance.''
    \begin{itemize}
        \item \texttt{The Egyptian Cat, also known as the Egyptian Mau, is characterized by its medium size, muscular build, distinctive spots on its fur, a black stripe running from its eyes to the neck, and green or gold eyes.}
        \item \texttt{The space bar has a long, rectangular shape with a concave top and a convex bottom, and is wider than most other keys on the keyboard.}
        \item \texttt{Hair spray usually comes in a cylindrical container with a long nozzle that sprays a fine mist, and may have a cap or cover on top to protect the nozzle.}
    \end{itemize}
\end{itemize}

\section{More details on Comparing Different Few Shot Learning Strategies}

In this section, we provide more implementation details of our ablation experiment in Sec.~\ref{sec:ablations}, where we compare finetuning student visual backbone vs. training-free retrieval for few shot learning on $\mathcal{X}_{\textrm{ood}}$. Finetuning student visual backbone follows the same settings in Sec.~\ref{sec:app_hyperparams}. For training-free retrieval, we adopt an approach similar to Tip-Adapter~\cite{tipadapter2022}, except that we use the student image model $S$ as the visual encoder and teacher's language model $T_{\textrm{img}}$ as the textual encoder. We set $\alpha=1.0$ and $\beta=5.5$.

\section{More details on Sec.~\ref{sec:applications} Application}
\label{sec:app_more_details_application}
For our experiments in Sec.~\ref{sec:applications}, we are in a multi-label classification setting, and the student network needs to output whether it is feasible to grasp each YCB object given the current observation. During student training, we find learning a common positive prompt and a common negative prompt for all labels to be very helpful, and we adopt DualCoOp~\cite{dualcoop2022} to learn these prompts. To encourage the global student visual feature be region-aware, we adopt the region aggregation method from DualCoOp for all of our experiments. We additionally calibrate the positive and negative probabilities (for each label, positive probability + negative probability = 1) by learning a common probability bias on the fly. An object $y$ is then predicted as positive if its positive probability is greater than $0.5 + \textrm{bias}$.

The two evaluation metrics we calculated in Sec.~\ref{sec:applications} can be formally defined as follows:
\begin{itemize}
    \item Overall accuracy over all YCB objects in the label set $\mathcal{Y}$ ($\mathcal{Y}=\mathcal{Y}_{\textrm{id}}$ on $\mathcal{X}_{\textrm{id}}$ and $\mathcal{Y}=\mathcal{Y}_{\textrm{id}} \cup \mathcal{Y}_{\textrm{ood}}$ on $\mathcal{X}_{\textrm{ood}}$): 
    \begin{equation}
        \mathcal{M}_1 = \frac{\sum_{i=1}^{|\mathcal{Y}|} \sum_{j=1}^{|\mathcal{X}|} \textrm{correct}(x_j, y_i) }{|\mathcal{X}||\mathcal{Y}|}
    \end{equation}
    Here $\textrm{correct}(x_j, y_i)$ outputs 1 if the existence of object $y_i$ is predicted correctly in the observation $x_j$, and 0 otherwise.
    \item F1-measure over objects that exist in an observation, averaged over all observations:
    \begin{equation}
    \begin{aligned}
        \mathcal{M}_{2, \textrm{precision}}(y) &= \frac{\sum_{i=1}^{|\mathcal{X}|} \mathbf{1}_{y \in \textrm{obj}(x_i)} * \textrm{correct}(x_i, y)}{\sum_{i=1}^{|\mathcal{X}|} \min(\mathbf{1}_{y \in \textrm{obj}(x_i)} + \textrm{pred}(x_i, y), 1)} \\
        \mathcal{M}_{2,\textrm{precision}} &= \frac{\sum_{y \in \mathcal{Y}} \mathcal{M}_{2, \textrm{precision}}(y)}{|\mathcal{Y}|} \\
        \mathcal{M}_{2, \textrm{recall}}(y) &= \frac{\sum_{i=1}^{|\mathcal{X}|} \mathbf{1}_{y \in \textrm{obj}(x_i)} * \textrm{correct}(x_i, y)}{\sum_{i=1}^{|\mathcal{X}|} \mathbf{1}_{y \in \textrm{obj}(x_i)}} \\
        \mathcal{M}_{2,\textrm{recall}} &= \frac{\sum_{y \in \mathcal{Y}} \mathcal{M}_{2, \textrm{recall}}(y)}{|\mathcal{Y}|} \\
        \mathcal{M}_{2,\textrm{F1}} &= \frac{2}{\frac{1}{\mathcal{M}_{2, \textrm{precision}}} + \frac{1}{\mathcal{M}_{2, \textrm{recall}}}}
    \end{aligned}
    \end{equation}
    Here $\textrm{pred}(x_i, y)$ equals 1 if the student network predicts that the object $y$ exists in observation $x_i$, and 0 otherwise.
\end{itemize}

\end{document}